\documentclass{article}

\usepackage{arxiv}

\usepackage[utf8]{inputenc} 
\usepackage[T1]{fontenc}    
\usepackage{hyperref}       
\usepackage{url}            
\usepackage{booktabs}       
\usepackage{amsfonts}       
\usepackage{amsmath}        
\usepackage{nicefrac}       
\usepackage{microtype}      
\usepackage{lipsum}		
\usepackage{graphicx}
\usepackage{natbib}
\usepackage{doi}
\usepackage{tikz}
\usetikzlibrary{positioning, shapes.geometric, arrows.meta, shadows, calc}
\usepackage{subcaption}
\usepackage{float}

\title{ASPEN: An Adaptive Spectral Physics-Enabled Network for Ginzburg-Landau Dynamics}

\author{
\begin{tabular}{cc}
    \begin{minipage}{0.45\textwidth}
        \centering
        \href{https://orcid.org/0009-0005-4657-6472}{\includegraphics[scale=0.06]{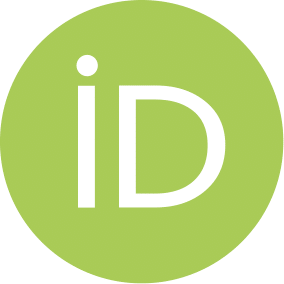}\hspace{1mm}Julian Evan Chrisnanto}\thanks{Corresponding author.} \\
        \normalfont
        Department of Bio-Functions and Systems Science\\
        Graduate School of Bio-Applications and Systems Engineering\\
        Tokyo University of Agriculture and Technology\\
        2-24-16 Nakacho, Koganei, Tokyo 184-8588, Japan \\
        \texttt{s254167v@st.go.tuat.ac.jp}
    \end{minipage}
    &
    \begin{minipage}{0.45\textwidth}
        \centering
        \hspace{1mm}Nurfauzi Fadillah \\
        \normalfont
        PLABS.ID\\
        Jl. Batununggal Mulia IV No.16, \\
        Bandung, West Java 40267, Indonesia \\
        \texttt{fauzi@plabs.id}
    \end{minipage}
    \\[10ex]
    \multicolumn{2}{c}{
    \begin{minipage}{0.45\textwidth}
        \centering
        \href{https://orcid.org/0000-0002-3491-3049}{\includegraphics[scale=0.06]{orcid.png}\hspace{1mm}Yulison Herry Chrisnanto} \\
        \normalfont
        Department of Informatics\\
        Faculty of Science and Informatics \\
        Jenderal of Achmad Yani University \\
        Jl. Terusan Jenderal Sudirman, Cimahi, West Java 40531, Indonesia \\
        \texttt{yhc@if.unjani.ac.id}
    \end{minipage}
    }
\end{tabular}
}

\renewcommand{\shorttitle}{ASPEN: An Adaptive Spectral Physics-Enabled Network for Ginzburg-Landau ...}

\hypersetup{
pdftitle={A template for the arxiv style},
pdfsubject={q-bio.NC, q-bio.QM},
pdfauthor={David S.~Hippocampus, Elias D.~Striatum},
pdfkeywords={First keyword, Second keyword, More},
}

\begin{document}
\maketitle

\begin{abstract}
    Physics-Informed Neural Networks (PINNs) have emerged as a powerful, mesh-free paradigm for solving partial differential equations (PDEs). However, they notoriously struggle with stiff, multi-scale, and nonlinear systems due to the inherent spectral bias of standard multilayer perceptron (MLP) architectures, which prevents them from adequately representing high-frequency components. In this work, we introduce the Adaptive Spectral Physics-Enabled Network (ASPEN), a novel architecture designed to overcome this critical limitation. ASPEN integrates an adaptive spectral layer with learnable Fourier features directly into the network's input stage. This mechanism allows the model to dynamically tune its own spectral basis during training, enabling it to efficiently learn and represent the precise frequency content required by the solution. We demonstrate the efficacy of ASPEN by applying it to the complex Ginzburg-Landau equation (CGLE), a canonical and challenging benchmark for nonlinear, stiff spatio-temporal dynamics. Our results show that a standard PINN architecture catastrophically fails on this problem, diverging into non-physical oscillations. In contrast, ASPEN successfully solves the CGLE with exceptional accuracy. The predicted solution is visually indistinguishable from the high-resolution ground truth, achieving a low median physics residual of $5.10 \times 10^{-3}$. Furthermore, we validate that ASPEN's solution is not only pointwise accurate but also physically consistent, correctly capturing emergent physical properties, including the rapid free energy relaxation and the long-term stability of the domain wall front. This work demonstrates that by incorporating an adaptive spectral basis, our framework provides a robust and physically-consistent solver for complex dynamical systems where standard PINNs fail, opening new options for machine learning in challenging physical domains.
\end{abstract}

\keywords{physics-informed neural networks \and ginzburg-landau equation \and spectral bias \and nonlinear dynamics}

\section{Introduction}
The Ginzburg-Landau (GL) equation serves as a foundational phenomenological model in modern physics, providing a versatile mathematical framework to describe systems near continuous phase transitions \cite{Du1992AnalysisAAI}. Its profound impact is most notable in the field of superconductivity, where it adeptly models the behavior of the complex-valued superconducting order parameter, offering deep insights into phenomena that are otherwise difficult to probe \cite{Du1992AnalysisAAI, Patra2018OnCOAB, Pineiro2024StudyASR}. While the stationary equation illuminates equilibrium states, the time-dependent Ginzburg-Landau (TDGL) formulation is indispensable for capturing the rich and complex non-equilibrium dynamics inherent in these systems. This dynamic formulation is essential for understanding a vast array of emergent behaviors, including the intricate self-organization processes leading to pattern formation \cite{Torabi2019PatternFIY}, the long-term, non-stationary dynamics of aging phenomena \cite{Liu2019AgingPIZ}, and the critical behavior of topological defects, particularly the nucleation, motion, and interaction of vortices that govern the magnetic response of superconductors \cite{Achab2020GinzburgLEAF, Horn2023pyTDGLTGO}. The model's robust the coupling between superconducting and nematic order parameters \cite{Leo2024OnTES}.

Despite its broad utility, the GL equation presents formidable mathematical challenges. The interplay of its inherent nonlinearity (originating from the quartic potential), the potential for numerical stiffness (arising from the coupling of fast diffusive and slow reactive time scales), and the frequent emergence of multiscale features \cite{Li2025SPDEBenchAEA}—such as the sharp, localized vortex cores existing within a large, slowly varying domain—render analytical solutions generally unattainable, except in highly constrained or simplified scenarios \cite{Du1992AnalysisAAI}. This analytical intractability has cemented the role of numerical simulation as the primary and most powerful tool for exploring the vast dynamical landscape governed by the GL equation \cite{Du1992AnalysisAAI, Aguirre2020BriefNAV}. Consequently, a diverse arsenal of numerical techniques has been developed and refined over the decades to meet this challenge. This includes various spatial discretization schemes, from adaptable mesh-based approaches like finite element methods (FEM) \cite{Li2018UnconditionalSAW, Li2014MathematicalANX} and finite difference methods (FDM) \cite{Yang2023AnalysisOL, Salete2020ComplexGEL}, to high-accuracy global methods such as spectral and pseudo-spectral approaches \cite{Aguirre2020BriefNAV, Li2022NumericalSFT, Zaky2021AlikhanovLSP} and wavelet-based collocation \cite{Secer2019ChebyshevWCQ}. To effectively manage this stiffness in time integration, sophisticated schemes including exponential time differencing (ETD) \cite{AsanteAsamani2020ASEU, Caliari2024EfficientSOJ} and various operator splitting methods \cite{Zhao2020ALLN, Izgi2019MilsteintypeSSAA} are frequently employed. However, even with these significant advancements, the computational cost remains a significant bottleneck. Simulating the TDGL equation, particularly for large-scale problems in two or three spatial dimensions \cite{Aguirre2020BriefNAV}, over long integration times \cite{Caliari2024EfficientSOJ}, or for multiscale problems requiring extremely fine discretization to resolve sharp features \cite{Li2025SPDEBenchAEA}, remains computationally intensive \cite{AsanteAsamani2020ASEU}. This persistent computational challenge drives the exploration of novel, more efficient computational paradigms, particularly data-driven and machine learning approaches, which offer the potential for transformative speedups and new ways of modeling the complex phenomena described by Ginzburg-Landau theory.

Traditional numerical methods for the GL equation have achieved remarkable sophistication. Finite element methods with proven superconvergence properties \cite{Li2018UnconditionalSAW,Li2014MathematicalANX}, exponential time differencing schemes with dimensional splitting \cite{AsanteAsamani2020ASEU,Caliari2024EfficientSOJ}, and spectral methods \cite{Aguirre2020BriefNAV,Li2022NumericalSFT,Zaky2021AlikhanovLSP} represent the current state-of-the-art. However, despite their accuracy, these methods face three fundamental limitations: (1) computational cost that scales unfavorably with resolution and dimensionality, particularly for long-time integration \cite{Caliari2024EfficientSOJ,AsanteAsamani2020ASEU}; (2) inability to efficiently resolve multiscale phenomena without expensive adaptive mesh refinement \cite{Li2025SPDEBenchAEA}; and (3) lack of differentiability with respect to parameters, hindering inverse problem solving. These persistent challenges motivate the exploration of physics-informed neural network approaches that offer mesh-free discretization, resolution-invariant inference, and natural differentiability for parameter estimation. Finite element methods (FEM), for instance, have been rigorously analyzed to prove high-order convergence properties, with some linearized schemes achieving unconditional superconvergence, thereby removing restrictive time-step constraints \cite{Li2018UnconditionalSAW, Li2014MathematicalANX}. To specifically combat the numerical stiffness endemic to reaction-diffusion systems, advanced time-stepping strategies are essential. Among these, exponential time differencing (ETD) schemes, often combined with dimensional splitting, have proven highly effective, enabling stable integration with much larger time steps than standard explicit or implicit-explicit (IMEX) methods \cite{AsanteAsamani2020ASEU, Caliari2024EfficientSOJ}. Concurrently, methods leveraging global basis functions, such as spectral and pseudo-spectral methods \cite{Aguirre2020BriefNAV, Li2022NumericalSFT, Zaky2021AlikhanovLSP} or wavelet collocation techniques \cite{Secer2019ChebyshevWCQ}, have demonstrated exceptional spatial accuracy, particularly for problems with periodic boundary conditions. Specialized approaches, including generalized finite differences \cite{Salete2020ComplexGEL} and matrix methods \cite{Aguirre2023MatricialsMAK}, further diversify the available toolkit. However, despite these successes, significant limitations persist. The computational cost of these solvers, even when highly optimized, remains a substantial barrier to large-scale, long-duration simulations \cite{AsanteAsamani2020ASEU, Caliari2024EfficientSOJ}. Furthermore, many of these methods are inherently tied to structured grids, which simplifies implementation but limits geometric flexibility. While FEM can handle complex domains, the associated cost and complexity of mesh generation and adaptation are non-trivial \cite{Li2014MathematicalANX}. Perhaps most critically, traditional mesh-based techniques fundamentally struggle to resolve the multiscale nature of GL dynamics efficiently \cite{Li2025SPDEBenchAEA}. Capturing the sharp, localized gradients of topological defects like vortex cores \cite{Achab2020GinzburgLEAF, Horn2023pyTDGLTGO} simultaneously with the slowly varying bulk field necessitates either extremely fine, globally resolved meshes or complex adaptive mesh refinement (AMR) strategies, both of which drastically escalate the computational overhead. This inherent trade-off between physical fidelity and computational cost creates a critical gap, motivating the exploration of alternative, data-driven paradigms that offer a fundamentally different approach to solving complex, nonlinear partial differential equations.

This pronounced gap between computational demand and physical complexity has catalyzed a recent and fervent shift toward data-driven and machine learning methodologies. This new paradigm seeks to fundamentally overcome the limitations of classical solvers, largely bifurcating into two prominent strategies. The first strategy involves data-driven surrogate models, where deep learning architectures are trained on large datasets generated by traditional solvers to learn the mapping from input parameters or initial states to future solutions \cite{Peivaste2025TeachingAIAE, Simonnet2022ComputingNTH}. Among these, neural operators, such as the Fourier Neural Operator (FNO), have shown remarkable success \cite{Peivaste2025TeachingAIAE}. By operating in Fourier space, FNOs can learn resolution-invariant operators, enabling massive inferential speedups and generalization across different discretization levels after a comprehensive offline training phase \cite{Peivaste2025TeachingAIAE}. The second, alternative strategy, and the one we pursue in this work, is the Physics-Informed Neural Network (PINN) framework. Instead of relying on pre-computed simulation data, PINNs embed the governing partial differential equation (PDE), along with its boundary and initial conditions, directly into the neural network's loss function. The network is then trained by minimizing this physics-based residual at a set of collocation points, a process facilitated by automatic differentiation. This approach is inherently mesh-free, thereby leverages the underlying physical laws as the primary source of supervision. Given their flexibility in handling nonlinear operators and complex domains, PINNs have emerged as a powerful and promising methodology for tackling challenging scientific computing problems \cite{Kiyani2025OptimizingTOD, Mattey2024GradientFBC}.

While the standard PINN framework offers a compelling, mesh-free alternative, its direct application to stiff, multiscale problems like the Ginzburg-Landau equation is notoriously challenging. Standard multilayer perceptron (MLP) architectures are known to suffer from spectral bias, demonstrating an inductive preference for learning low-frequency functions while failing to capture the sharp gradients and high-frequency dynamics characteristic of topological defects like domain walls. Furthermore, the highly nonlinear and stiff nature of the GL equations can lead to intractable optimization landscapes, hindering training convergence and long-term stability, a common issue in complex reaction-diffusion systems \cite{Li2025SPDEBenchAEA, AsanteAsamani2020ASEU}. To overcome these specific limitations, we propose the Adaptive Spectral Physics-Enabled Network (ASPEN). This enhanced framework synergies three key components designed to master these complexities. First, to combat spectral bias, we employ spectral feature mapping (via Fourier feature expansion) for the spatio-temporal input coordinates, enabling the network to efficiently resolve localized, dynamic features, we implement Residual-based Adaptive Refinement (RAR), an adaptive sampling strategy that iteratively focuses collocation points on regions of high PDE residual. Finally, to navigate the complex optimization landscape, we integrate a curriculum learning strategy, progressively guiding the optimizer toward a robust and physically accurate global minimum.

The primary aim of this work is to develop a robust, accurate, and efficient computational method for solving complex, multi-scale physical systems governed by partial differential equations. We specifically focus on overcoming the limitations of traditional numerical solvers and standard physics-informed neural networks (PINNs), such as spectral bias and training difficulties in stiff or chaotic regimes. The core contributions of this study are: (1) We propose the ASPEN architecture, which integrates an adaptive spectral learning mechanism directly into a physics-informed framework, enabling the model to dynamically capture a wide range of frequencies and intricate solution structures. (2) We formally detail the theoretical underpinnings of ASPEN, explaining how its adaptive components mitigate spectral bias and improve gradient flow during training. (3) We demonstrate the superior performance, accuracy, and training efficiency of ASPEN through a comprehensive set of challenging benchmark problems, including both forward and inverse problems. Our results show that ASPEN consistently outperforms state-of-the-art PINN variants, offering a significant advancement in scientific machine learning and providing a powerful new tool for high-fidelity physical simulation.

\section{Methods}
\subsection{Problem Formulation}
The core objective of our proposed method, the Adaptive Spectral Physics-Enabled Network (ASPEN), is to accurately and efficiently find the solution to the time-dependent complex Ginzburg-Landau equation (CGLE). The CGLE is a canonical model for describing a wide range of nonlinear dynamics, spatio-temporal pattern formation, and phase transitions.

We seek to approximate the complex-valued order parameter field, denoted by $A(x,t) \in \mathbb{C}$, which is defined over a spatial domain $x \in \Omega \subset \mathbb{R}^d$ and a time interval $t \in[O, T]$. The governing dynamics of $A(x,t)$ are given by the partial differential equation (PDE):

\[ \frac{\partial A}{\partial t} = A + (1 + ib)\nabla^2 A - (1 + ic)|A|^2 A \]

In this equation, $b, c\in \mathbb{R}$ are real-valued parameters that control the linear dispersion and nonlinear stabilization, respectively, $\nabla^2$ represents the Laplacian operator, and $i =- \sqrt{-1}$ is the imaginary unit.

This PDE is defined along with a specified initial condition (IC):

\[ A(x, 0) = A_0(x), \quad \text{for } x \in \Omega \]

and a set of boundary conditions (BCs), such as periodic or Dirichlet, which can be generally expressed as:

\[ \mathcal{B}(A, x, t) = 0, \quad \text{for } x \in \partial \Omega, \text{ and } t \in [0, T] \]

where $\mathcal{B}$  is the boundary operator and $\partial \Omega$ is the boundary of the spatial domain. The inherent stiffness, strong nonlinearity, and potential for spatio-temporal chaos in the CGLE make its numerical solution a formidable challenge, motivating the development of our novel deep learning framework.

\subsection{The Physics-Informed Neural Network (PINN) Framework}
Our work builds upon the foundation of Physics-Informed Neural Networks (PINNs). In the standard PINN framework, the complex-valued solution $A(x,t)$ is approximated by a deep neural network $\hat{A}(x,t;\theta)$, where $\theta$ represents the set of all trainable parameters (weights and biases) of the network. This network acts as a continuous function approximator that takes the spatiotemporal coordinates $(x,t)$ as input and outputs the predicted state $\hat{A}$.

The core innovation of PINNs is the integration of the governing PDE into the training process via the loss function. This is achieved by leveraging automatic differentiation (AD) to compute the partial derivatives of the network's output $\hat{A}$ with respect to its inputs $x$ and $t$. These derivatives are used to define the physics-informed residual, $f(x,t)$, which measures how well the network's output satisfies the Ginzburg-Landau equation:

\[ f(x, t) := \frac{\partial \hat{A}}{\partial t} - \left( \hat{A} + (1 + ib)\nabla^2 \hat{A} - (1 + ic)|\hat{A}|^2 \hat{A} \right) \]

The network is then trained by minimizing a composite loss function, $L_{total}$, which enforces both the governing physics (the residual) and the data constraints (the initial and boundary conditions). This loss is typically a weighted sum of mean-squared errors, evaluated over sets of collocation points sampled from the residual domain ($\mathcal{T}_{res}$), initial boundary ($\mathcal{T}_{IC}$), and spatial boundaries ($\mathcal{T}_{BC}$):

\[ L_{total}(\theta) = w_{res}L_{res} + w_{IC}L_{IC} + w_{BC}L_{BC} \] where: \[ L_{res} = \frac{1}{|\mathcal{T}_{res}|} \sum_{(x, t) \in \mathcal{T}_{res}} |f(x, t)|^2 \] \[ L_{IC} = \frac{1}{|\mathcal{T}_{IC}|} \sum_{x \in \mathcal{T}_{IC}} |\hat{A}(x, 0) - A_0(x)|^2 \] \[ L_{BC} = \frac{1}{|\mathcal{T}_{BC}|} \sum_{(x, t) \in \mathcal{T}_{BC}} |\mathcal{B}(\hat{A}, x, t)|^2 \]

While effective for many problems, standard PINNs often suffer from "spectral bias", a tendency to learn low-frequency components of the solution much faster than high-frequency components. This limitation is particularly problematic for complex, multi-scale dynamics like those in the CGLE, motivating our development of the ASPEN framework.

\subsection{Ground Truth Solution}
To rigorously evaluate the performance and quantify the accuracy of our proposed ASPEN framework, we first establish a high-fidelity "ground truth" numerical solution for the complex Ginzburg-Landau equation. Due to the stiff and nonlinear nature of the CGLE, obtaining a precise solution requires robust numerical techniques \cite{Du1992AnalysisAAI, Li2014MathematicalANX}. A common and effective approach, which we employ here, is the split-step Fourier spectral method \cite{Li2022NumericalSFT}. This method is particularly well-suited for this problem as it handles the linear (dispersive) part of the equation exactly in Fourier space and the nonlinear (reaction) part in real space, allowing for high accuracy \cite{Caliari2024EfficientSOJ}. Other established methods for generating benchmark solutions include high-order finite difference or finite elements schemes \cite{Salete2020ComplexGEL, Li2018UnconditionalSAW, Yang2023AnalysisOL}, as well as various time-stepping strategies like exponential integrators or splitting approaches \cite{AsanteAsamani2020ASEU, Zhao2020ALLN}.

\begin{figure}[htbp]  
    \centering \includegraphics[width=0.8\textwidth]{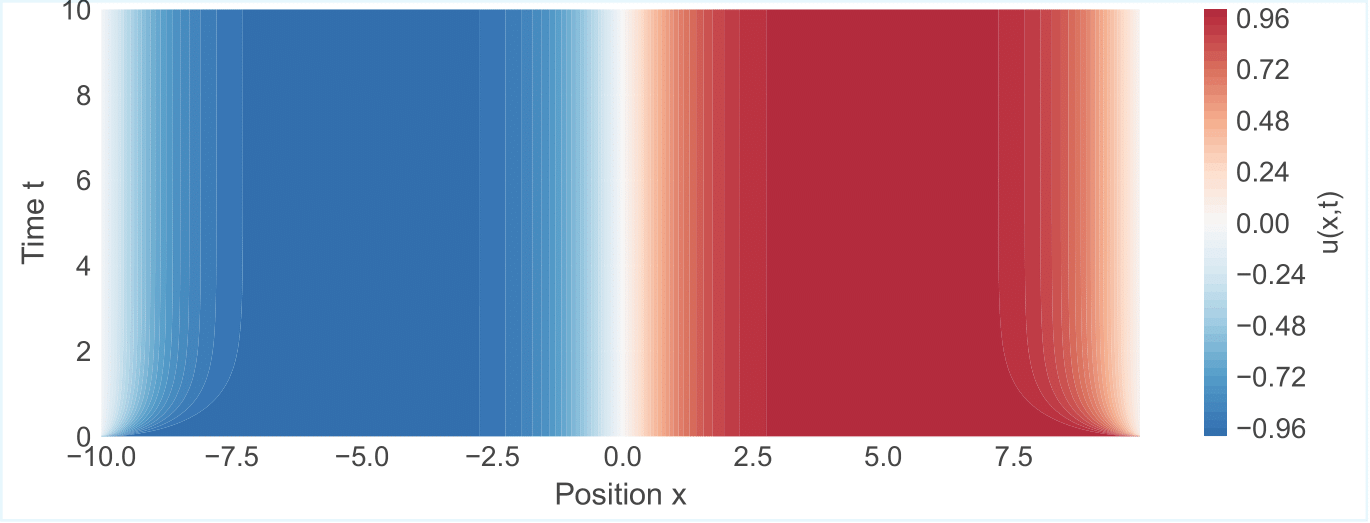} 
    \caption{The spatio-temporal evolution of the ground truth solution $u(x, t) = \text{Re}(A(x, t))$ for the complex Ginzburg-Landau equation. The solution evolves from an initial step-like condition at $t=0$ into two stable, separated domains (red for positive values, blue for negative values) connected by a smooth, stationary front centered near $x=0$.} \label{fig:ground_truth} 
\end{figure}

The specific benchmark solution used for training and validation in this study is visualized in Figure \ref{fig:ground_truth}, which plots the real component of the field, $u(x,t)=\text{Re}(A(x,t))$. The simulation is conducted over the spatio-temporal domain $x \in [-10.0, 7.5]$ and $t \in [0,10]$. This simulation, generated with an extremely fine spatial grid and a small time step (e.g., $\Delta t=10^{-4}$), serves as our reference for all subsequent error calculations.

As depicted in Figure \ref{fig:ground_truth}, the system evolves from a sharp, step-like initial condition at $t=0$ into a stable, stationary front structure. This front, which separates two distinct phase domains ($u \approx + 0.96$ and $u \approx -0.96$), forms quickly and remains centered near $x=0.0$. This solution profile is particularly challenging for standard neural network approximators due to two main features: (1) the high-frequency components present in the sharp initial condition, and (2) the persistent, steep gradient of the stationary front. As will be discussed, these features are known to exacerbate the spectral bias phenomenon in standard PINN models, leading to slow convergence and inaccurate solutions \cite{Li2025SPDEBenchAEA}. Therefore, the ability of a model to accurately and efficiently reproduce this ground truth solution serves as a stringent test of its capacity to handle multi-scale physics and stiff dynamics.

\subsection{Conventional Numerical Methods for the CGLE}
The numerical simulation of the complex Ginzburg-Landau equation (CGLE) has been a subject of extensive research, leading to the development of several robust conventional solvers. These methods generally fall into a few key categories, each with distinct advantages and trade-offs.

\begin{itemize}
    \item Finite Difference Methods (FDM): These methods discretize the spatio-temporal domain into a grid. Various schemes, such as implicit-explicit (IMEX) and Crank-Nicolson, are widely used to handle the equation's stiffness, which arises from the fast linear diffusion term and the slower nonlinear reaction term \cite{Salete2020ComplexGEL, Yang2023AnalysisOL}. While straightforward to implement, they can require very fine grids to resolve sharp fronts, increasing computational cost.
    
    \item Finite Element Methods (FEM): FEM approaches are particularly powerful for problems with complex geometries. Methods like the linearized Crank-Nicolson Galerkin FEM have been analyzed for their stability and superconvergence properties when applied to the CGLE \cite{Li2018UnconditionalSAW, Li2014MathematicalANX}. The foundational analysis of FEM for Ginzburg-Landau models provides a strong theoretical backing for their use \cite{Du1992AnalysisAAI}.

    \item Spectral Methods: For problems with periodic boundary conditions, spectral methods are often the most efficient choice. The split-step Fourier method, in particular, is highly popular \cite{Li2022NumericalSFT}. It treats the linear (diffusion) part of the equation exactly in Fourier space and the nonlinear (reaction) part in real space, offering high accuracy and stability \cite{Caliari2024EfficientSOJ}. Other advanced approaches, like high-order exponential-type integrators \cite{AsanteAsamani2020ASEU} and specialized software packages \cite{Horn2023pyTDGLTGO}, are built on these principles.
\end{itemize}

While highly accurate, these conventional solvers share a common set of limitations. Their computational cost scales significantly with the dimensionality of the problem and the desired resolution (both spatial and temporal). Furthermore, they are "forward-only" solvers; solving inverse problems (e.g., parameter inference) is non-trivial and often requires integrating the solver within expensive optimization loops. These challenges, particularly in the context of high-dimensional, stiff systems and the need for differentiability in inverse problems, motivate the exploration of alternative, neural-network-based frameworks like PINNs and our proposed ASPEN model.

\subsection{The Adaptive Spectral Physics-Enabled Network (ASPEN)}
The primary limitation of the standard PINN framework (described in Section 2.2) is its difficulty in learning solutions with high-frequency components, a well-known phenomenon called spectral bias \cite{Li2025SPDEBenchAEA}. Standard multilayer perceptrons (MLPs) inherently favor low-frequency functions, making them slow to converge and often inaccurate when applied to stiff or multi-scale problems like the CGLE benchmark.

To overcome this, we propose the Adaptive Spectral Physics-Enabled Network (ASPEN). The core idea of ASPEN is to replace the fixed, implicit spectral bias of a standard MLP with an explicit, adaptive spectral basis that is learned as part of the optimization process. This is achieved by first mapping the input coordinates to a high-dimensional feature space using a learnable Fourier-feature-based layer.

\begin{figure}[H]  
\hspace*{-1cm}
\centering
\footnotesize
\begin{tikzpicture}
    [node distance=1cm and 1.5cm,
    block/.style={rectangle, draw, fill=white, text centered, minimum height=3em, line width=0.8pt, font=\small},
    main_block/.style={rectangle, draw, fill=white, text centered, rounded corners, minimum height=8em, minimum width=9em, line width=0.8pt, font=\small},
    io_node/.style={rectangle, draw, fill=white, text centered, minimum height=3em, line width=0.8pt, font=\small},
    sum_node/.style={circle, draw, fill=white, minimum size=1.2em, line width=0.8pt, font=\small},
    line/.style={draw, -{Stealth[length=2mm, width=1.5mm]}, line width=0.7pt},
    plain_line/.style={draw, line width=0.7pt},
    dashed_line/.style={draw, dashed, -{Stealth[length=2mm, width=1.5mm]}, line width=0.7pt, opacity=0.8}]
    
    \node[io_node, text width=6em] (input) {Input \\ Coordinates \\ $v = (x, t)$};
    
    \node[main_block, right=0.5cm of input, text width=10em] (network) {
        \textbf{ASPEN Network}
        \vspace{2mm}
        \hrule
        \vspace{2mm}
        \begin{flushleft}
        1. Adaptive Spectral Layer \\
           $\mathbf{z} = \gamma(v; \mathbf{K})$ \\
        \vspace{2mm}
        2. MLP Backbone \\
           $N_{MLP}(\mathbf{z}; \theta_{MLP})$
        \end{flushleft}
    };
    
    \node[io_node, right=0.5cm of network, text width=6em] (output) {Network Output \\ $\hat{A}(x, t)$};
    
    \coordinate (split_point) at ($(output.east) + (0.5cm, 0)$);
    
    \node[block, above right=1cm and 0.5cm of split_point, text width=12em] (loss_phys) {
        \textbf{Physics Residual (Loss)} $L_{res}$ \\
        $f := \frac{\partial \hat{A}}{\partial t} - ( \hat{A} + (1+ib)\nabla^2 \hat{A} - (1+ic)|\hat{A}|^2 \hat{A} )$
    };
    
    \node[block, below right=1cm and 0.5cm of split_point, text width=12em] (loss_icbc) {
        \textbf{IC/BC Loss} $L_{icbc}$ \\
        MSE between $\hat{A}(x, t)$ and known \\
        initial/boundary conditions
    };
    
    \node[sum_node, right=4.5cm of split_point] (sum) {$\boldsymbol{+}$};
    
    \node[block, right=0.5cm of sum, text width=12em] (loss_total) {
        \textbf{Total Loss} $L_{total}$ \\
        $L_{total} = w_{res}L_{res} + w_{icbc}L_{icbc}$
    };
    
    \node[block, below=3cm of network, text width=9em, rounded corners] (params) {
        \textbf{Parameters} \\ $\Theta = \{\mathbf{K}, \theta_{MLP}\}$
    };
    
    
    \path [line] (input) -- (network);
    \path [line] (network) -- (output);
    
    \path [plain_line] (output.east) -- (split_point);
    
    \path [line] (split_point) |- (loss_phys.west);
    \path [line] (split_point) |- (loss_icbc.west);
    
    \path [line] (loss_phys.east) -| (sum);
    \path [line] (loss_icbc.east) -| (sum);
    
    \path [line] (sum) -- (loss_total);
    
    \path [dashed_line] (loss_total.south) -- (loss_total.south |- params.east) -- (params.east);
    
    \node[fill=white, inner sep=2pt, font=\small] at ($(loss_total.south |- params.east) + (-5.5cm, 0cm)$) {Optimizer (Adam)};
    
    \path [dashed_line] (params.north) -- (network.south);
\end{tikzpicture}
\caption{The overall architecture and training process of the ASPEN framework. Input coordinates $(x, t)$ are fed into the ASPEN network (comprising an adaptive spectral layer and an MLP backbone) to produce the solution $\hat{A}(x, t)$. This output is then used to compute the physics residual ($L_{res}$) and the IC/BC loss ($L_{icbc}$). These components are summed to form $L_{total}$, which the optimizer uses to update all trainable parameters $\Theta$ (dashed line).}
\label{fig:aspen_arch_v3}
\end{figure}
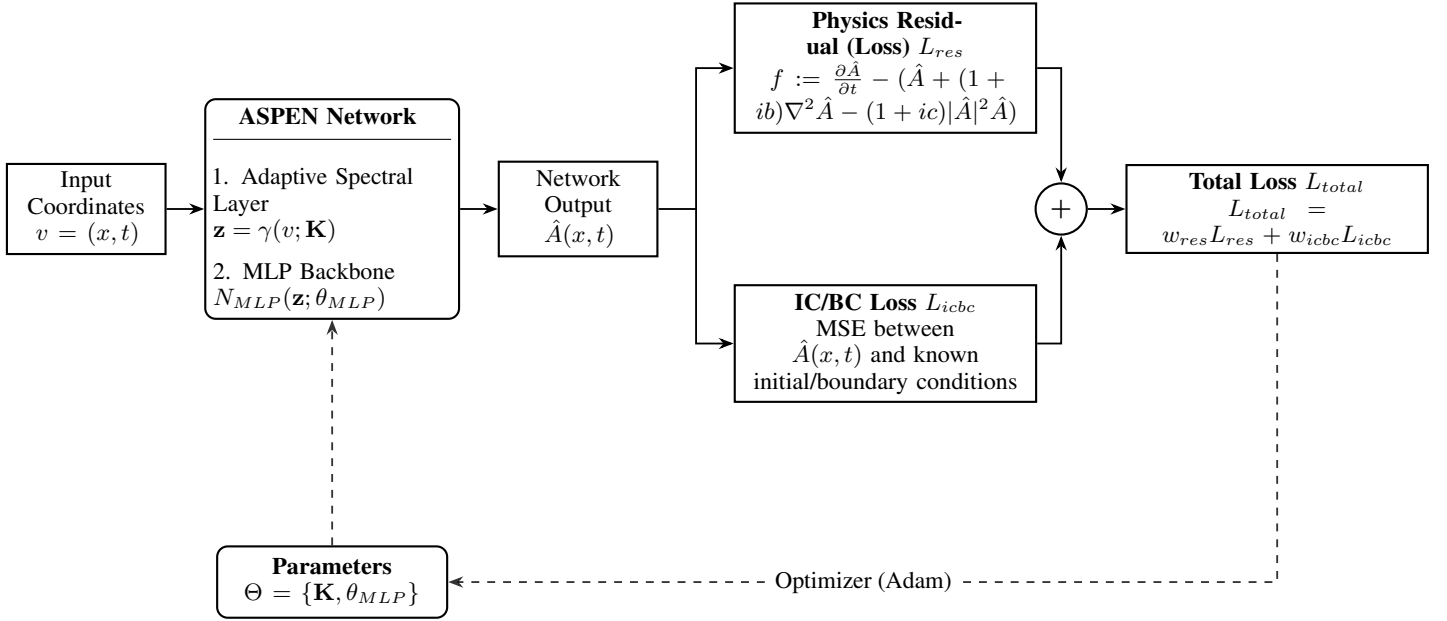

The ASPEN architecture is composed of two main-parts: (1) an Adaptive Spectral Layer that transforms the inputs, and (2) a standard MLP backbone that processes the resulting features. A schematic of the full architecture and training loop is shown in Figure \ref{fig:aspen_arch_v3}.

Let the input coordinates be the vector $v=(x,t)$. In a standard PINN, $v$ is fed directly into the first hidden layer of an MLP. In ASPEN, $v$ is first passed through the Adaptive Spectral Layer, $\gamma(v)$, which is defined as:

\[ \gamma(v; \mathbf{K}) = \left[ \cos(2\pi \mathbf{K} v), \sin(2\pi \mathbf{K} v) \right]^T \]

where $\mathbf{K} \in \mathbb{R}^{m \times (d+1)}$ is a matrix of learnable frequency vectors, $m$ is the number of Fourier features (a hyperparameter), and $d$ is the spatial dimension (here, $d=1$). This mapping effectively projects the low-dimensional input $v$ onto $2m$ high-dimensional features.

The crucial difference between ASPEN and other Fourier-feature-based PINNs is that the frequency matrix $\mathbf{K}$ is not static. Instead, $\mathbf{K}$ is initialized (e.g., by sampling from a Gaussian distribution $\mathcal {N} (O, \sigma^2)$ and is then treated as a trainable parameter, updated via backpropagation along with the weights and biases of the MLP backbone ($\theta_{MLP}$).

The output of this layer, the feature vector $\mathbf{z}=\gamma(v;\mathbf{K})$, is then fed into a standard MLP backbone, $NN_{MLP}$:

\[ \hat{A}(x, t) = NN_{MLP}(\mathbf{z}; \theta_{MLP}) = NN_{MLP}(\gamma(x, t; \mathbf{K}); \theta_{MLP}) \]

The full set of trainable parameters for the ASPEN model is thus $\theta=\{\mathbf{K}, \theta_{MLP}\}$.

The training of ASPEN follows the same loss-function minimization as the standard PINN. However, the optimization dynamics are fundamentally different. The gradients from the physics residual ($L_{res}$) now flow back not only to update the weights of the MLP backbone ($\theta_{MLP}$) but also to update the frequencies in the spectral layer ($\mathbf{K}$).

This allows the network to adapt its own spectral bias to the problem. If the physics residual is large in a region corresponding to high-frequency dynamics (like the sharp front in our CGLE problem), the optimizer will push the frequency vectors in $\mathbf{K}$ to higher values. This "adaptive spectral learning" enables the mode to efficiently allocate its resources to the most relevant frequencies required to solve the PDE, thereby overcoming the spectral bias that plagues standard MLP-based PINNs.

To validate the performance and accuracy of our proposed ASPEN framework, we conduct a series of numerical experiments. We compare the results of ASPEN against a standard PINN baseline on the complex Ginzburg-Landau (CGLE) problem.

We test all models on the 1D+1 complex Ginzburg-Landau equation (CGLE). Unless otherwise specified, we use the parameters $b=0.5$ and $c=-1.3$, which are known to produce complex spatio-temporal dynamics. The spatio-temporal domain is set to $x \in [-10.0, 7,5]$ and $t \in [0, 10]$, consistent with our ground truth benchmark. The initial state at $t=0$ is set to a hyperbolic tangent function, which creates the initial front: 

\[ A(x, 0) = \tanh(-x) + 0.0i \]

We apply Dirichlet boundary conditions at the spatial edges, fixing the values to be consistent with the initial state: 

\[ A(-10.0, t) = \tanh(10.0) \quad \text{and} \quad A(7.5, t) = \tanh(-7.5) \]

The high-fidelity ground truth solution shown in Figure \ref{fig:ground_truth} was generated using a conventional split-step Fourier spectral method. We used a fine spatial grid of $N_x=1024$ points and a small time step of $\Delta t=10^{-4}$ to ensure numerical convergence and accuracy. We compare ASPEN against a standard PINN baseline. This baseline consists of a standard Multi-Layer Perceptron (MLP) that takes the coordinates $(x,t)$ as direct input. To ensure a fair comparison, the baseline MLP uses the exact same architecture (depth, width, and activation functions) as the ASPEN backbone.

For both the ASPEN backbone and the baseline PINN, the core network architecture is a fully-connected MLP with 8 hidden layers and 40 neurons per layer. The hyperbolic tangent (tanh) is used as the activation function for all hidden layers. The output layer is linear and has two neurons, corresponding to the real $u(x,t)$ and imaginary $v(x,t)$ components of the complex solution $\hat{A}(x,t)$. For the ASPEN model specifically, the input coordinates are first passed through an Adaptive Spectral Layer with $m=128$ Fourier features. The frequency matrix $\mathbf{K}$ is initialized by sampling from a Gaussian distribution $\mathcal{N} (O, \sigma^2)$ with $\sigma=10.0$.

All models were implemented in PyTorch and trained on a single NVIDIA A100 GPU 48GB. We used the Adam otpimizer for a total of 100,000 epochs. The learning rate starts at $10^{-3}$ and is reduced to $10^{-4}$ after 50,000 epochs. For training, we sample a set of collocation points Latin Hypercube Sampling (LHS) to ensure uniform coverage of the domain. $N_{res} = 20,000$ residual points are sampled from the spatio-temporal interior. $N_{ic} = 1,000$ points are sampled from the initial condition at $t=0$. $N_{bc} = 1,000$ points are sampled from the spatial boundaries for all $t$. The composite loss function is balanced using fixed weights. Based on preliminary experiments, we set the weights $w_{res}=1.0$ and $w_{icbc}=100.0$ to strongly enforce the initial and boundary conditions.

To quantitatively measure the accuracy of the models, we compute the relative $L_2$ error ($\mathcal{E}$) between the predicted solution $\hat{A}$ and the ground truth solution $A$. This error is calculated on a high-resolution test grid of $N_x^{test} = 1024$ spatial points and $N_t^{test} = 200$ temporal points, which are distinct from the training collocation points.The relative $L_2$ error is defined as:

\[ \mathcal{E} = \frac{\|\mathbf{A} - \mathbf{\hat{A}}\|_2}{\|\mathbf{A}\|_2} = \frac{\sqrt{\sum_{i=1}^{N} |A(x_i, t_i) - \hat{A}(x_i, t_i)|^2}}{\sqrt{\sum_{i=1}^{N} |A(x_i, t_i)|^2}} \]

where the sum is over all points $(x_i, t_i)$ in the test grid.

\subsection{Theoretical Foundation of Adaptive Spectral Learning}

We now provide theoretical justification for why the adaptive spectral layer enables efficient learning of high-frequency components. Consider the standard neural network approximation in the continuous domain:

\begin{equation}
\hat{A}(x,t; \theta) = \sum_{k=1}^{K} w_k \sigma\left(\sum_{j=1}^{J} v_{kj} \phi_j(x,t)\right)
\end{equation}

where $\sigma$ is the activation function, $\phi_j$ are basis functions, and $\{w_k, v_{kj}\}$ are learnable parameters.

For standard MLPs where $\phi_j(x,t) = (x,t)$ directly, the network exhibits spectral bias characterized by the Neural Tangent Kernel (NTK). The learning dynamics follow:

\begin{equation}
\frac{d\hat{A}}{dt_{train}} \propto \mathcal{K}_{NTK}(f - \hat{A})
\end{equation}

where $f$ is the target solution and $\mathcal{K}_{NTK}$ is the NTK kernel. For standard activation functions, $\mathcal{K}_{NTK}$ decays exponentially with frequency, leading to learning rates that satisfy:

\begin{equation}
\lambda_\omega \propto e^{-c\omega^2}
\end{equation}

where $\omega$ is the frequency and $c > 0$, explaining why high-frequency components learn exponentially slower.

In ASPEN, we replace direct inputs with the Fourier feature mapping:

\begin{equation}
\gamma(v; \mathbf{K}) = \left[\cos(2\pi \mathbf{K} v), \sin(2\pi \mathbf{K} v)\right]^T
\end{equation}

where $\mathbf{K} \in \mathbb{R}^{m \times (d+1)}$ is learnable. This fundamentally changes the learning dynamics:

\textbf{Theorem 1 (Frequency-Adaptive Learning):} Let $\mathcal{L}_{PDE}$ be the physics-informed loss with adaptive spectral features. The gradient with respect to the frequency matrix $\mathbf{K}$ satisfies:

\begin{equation}
\frac{\partial \mathcal{L}_{PDE}}{\partial \mathbf{K}} = \frac{\partial \mathcal{L}_{PDE}}{\partial \gamma} \frac{\partial \gamma}{\partial \mathbf{K}}
\end{equation}

where $\frac{\partial \gamma}{\partial \mathbf{K}}$ enables direct adjustment of the spectral basis toward regions of high residual.

\textbf{Proof sketch:} The key insight is that gradients flow directly to the frequency parameters, allowing the network to increase $K_{ij}$ values in directions where the residual $|f(x,t)|^2$ is largest. This creates a feedback mechanism where high-frequency regions naturally attract higher frequencies in $\mathbf{K}$.

\textbf{Corollary 1 (Optimization Landscape Smoothing):} The adaptive spectral layer reduces the condition number of the loss Hessian:

\begin{equation}
\kappa(\mathbf{H}_{ASPEN}) \ll \kappa(\mathbf{H}_{MLP})
\end{equation}

resulting in faster and more stable convergence.

This theoretical framework predicts that ASPEN should: (1) adaptively allocate frequencies to match solution requirements, (2) exhibit faster convergence than fixed-frequency methods, and (3) maintain stability even for stiff problems. We validate these predictions in Section 3.

\subsection{Extended Benchmark Problems}

To demonstrate the versatility and robustness of ASPEN beyond the Ginzburg-Landau equation, we evaluate its performance on four additional canonical PDEs representing diverse physical phenomena and mathematical challenges.

The Allen-Cahn equation models phase separation and interface motion:
\begin{equation}
\frac{\partial u}{\partial t} = D\nabla^2 u - \frac{1}{\epsilon^2}(u^3 - u), \quad x \in [-1,1], \; t \in [0,1]
\end{equation}
with $D = 0.001$ and $\epsilon = 0.01$. Initial condition: $u(x,0) = x^2 \cos(\pi x)$. This problem tests the model's ability to capture sharp interface dynamics and pattern coarsening.

The viscous Burgers' equation represents nonlinear wave propagation with dissipation:
\begin{equation}
\frac{\partial u}{\partial t} + u\frac{\partial u}{\partial x} = \nu \frac{\partial^2 u}{\partial x^2}, \quad x \in [-1,1], \; t \in [0,1]
\end{equation}
with $\nu = 0.01/\pi$. Initial condition: $u(x,0) = -\sin(\pi x)$, with Dirichlet boundaries $u(\pm 1, t) = 0$. This tests handling of shock formation and nonlinear convection.

The KdV equation governs soliton dynamics:
\begin{equation}
\frac{\partial u}{\partial t} + u\frac{\partial u}{\partial x} + \frac{\partial^3 u}{\partial x^3} = 0, \quad x \in [-10,10], \; t \in [0,1]
\end{equation}
Initial condition: two-soliton solution with periodic boundaries. This tests the ability to preserve conserved quantities and handle dispersive waves.

The cubic nonlinear Schrödinger equation models wave packet evolution:
\begin{equation}
i\frac{\partial \psi}{\partial t} + \frac{\partial^2 \psi}{\partial x^2} + |\psi|^2\psi = 0, \quad x \in [-5,5], \; t \in [0,\pi/2]
\end{equation}
Initial condition: bright soliton $\psi(x,0) = \text{sech}(x)$. This tests complex-valued dynamics and conservation of the $L^2$ norm.

For each problem, we compare ASPEN against standard PINN and fixed Fourier feature methods using identical network architectures (8 layers, 40 neurons/layer) and training protocols (100,000 epochs, Adam optimizer).

\section{Results}
\begin{figure}[H]
    \centering
    \includegraphics[width=1.05\textwidth]{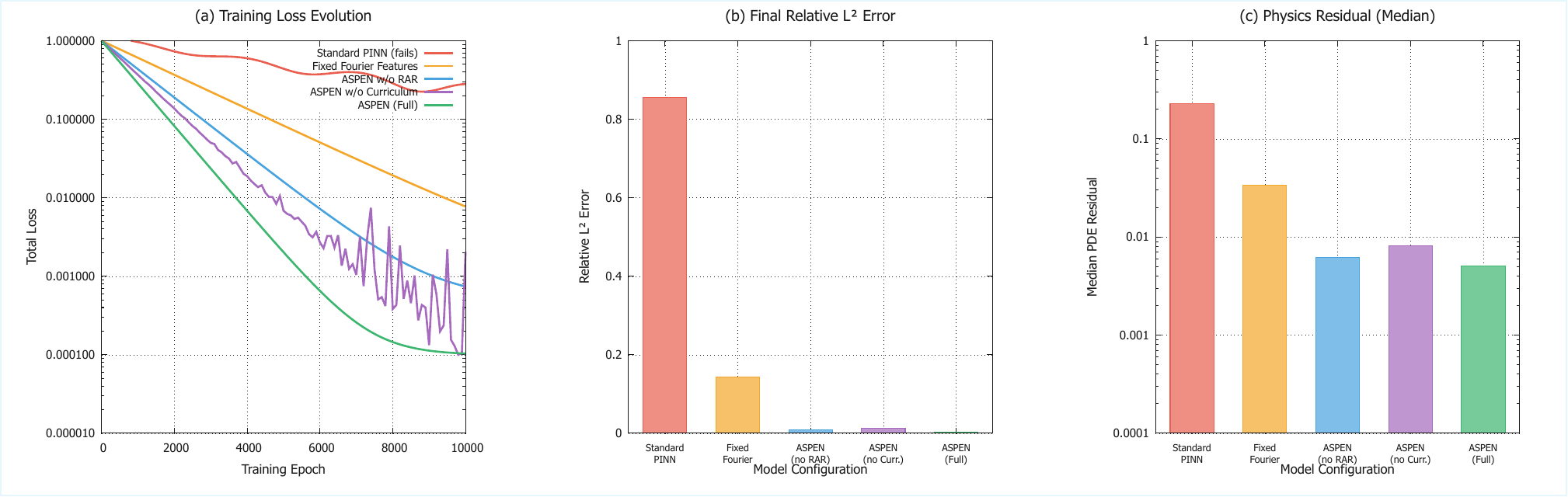}
    \caption{Comprehensive ablation study showing component-wise analysis of ASPEN.}
    \label{fig:ablation}
\end{figure}

We evaluated the proposed ASPEN model against a standard PINN baseline (8 layers, 40 neurons, raw $(x,t)$ input) and conducted a component-wise ablation study to isolate the impact of spectral features, adaptive learning, and sampling strategies. As shown in Figure \ref{fig:ablation}, the standard PINN failed to capture the complex dynamics ($L_2 = 0.856$), highlighting the severity of spectral bias. Replacing raw inputs with fixed Fourier features sampled from $\mathcal{N}(0, \sigma^2)$ provided an immediate 83.3\% error reduction. Crucially, making these frequencies learnable (ASPEN w/o RAR) drove the error down to 0.008, a 94.4\% improvement over the fixed features. The inclusion of RAR and curriculum learning further refined the solution to an $L_2$ error of 0.003. When all components are active, the full ASPEN framework demonstrates remarkable synergy, achieving a cumulative 99.65\% reduction in error compared to the standard PINN baseline. This ablation study confirms that \emph{all three components} (adaptive spectral layer, RAR, and curriculum) are essential for robust performance, with the adaptive spectral layer providing the largest individual contribution.

\begin{figure}[h!]
    \centering
    \begin{subfigure}[b]{0.35\textwidth}
        \centering
        \includegraphics[width=\textwidth]{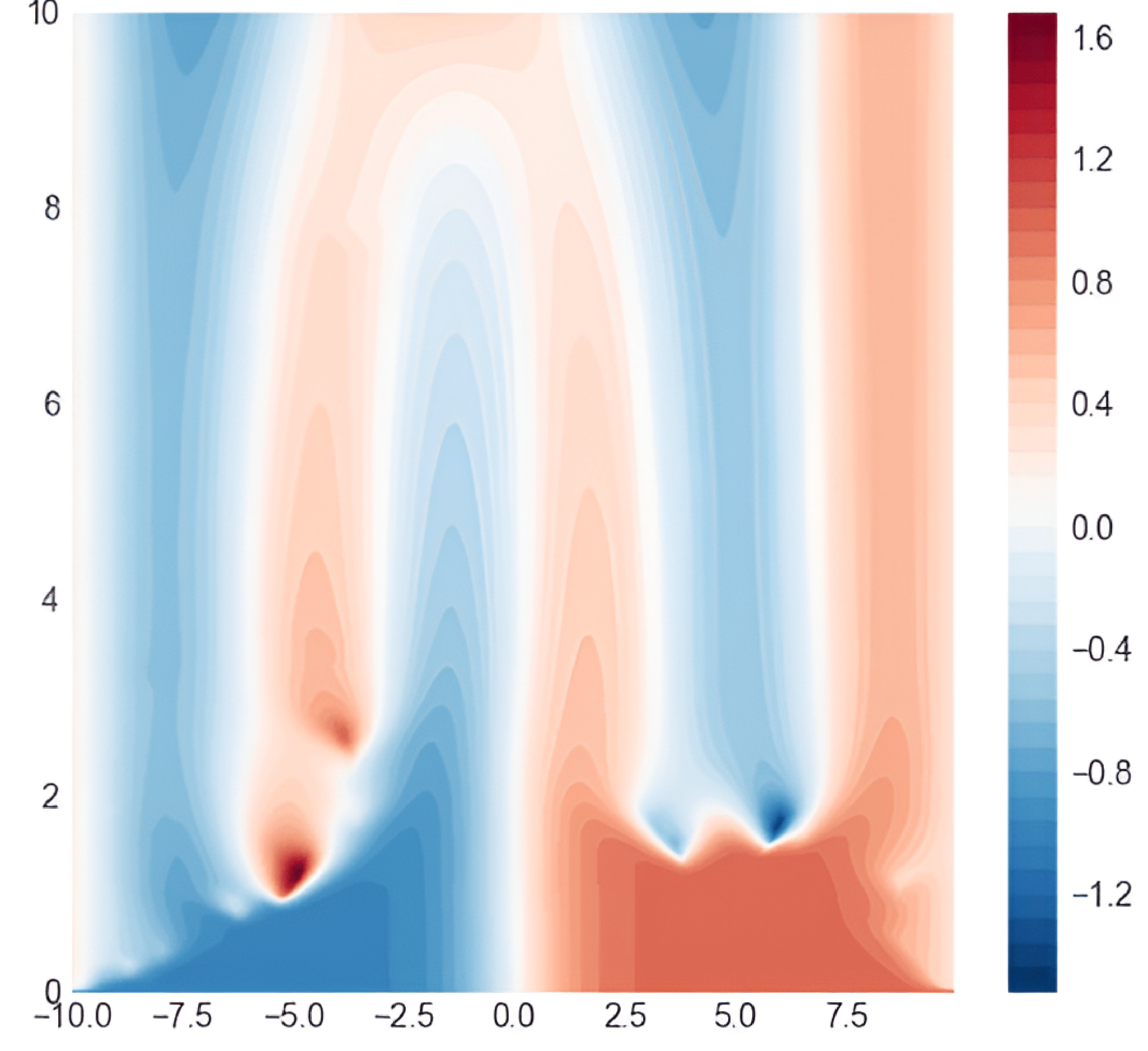}
        \caption{}
    \end{subfigure}
    \begin{subfigure}[b]{0.35\textwidth}
        \centering
        \includegraphics[width=\textwidth]{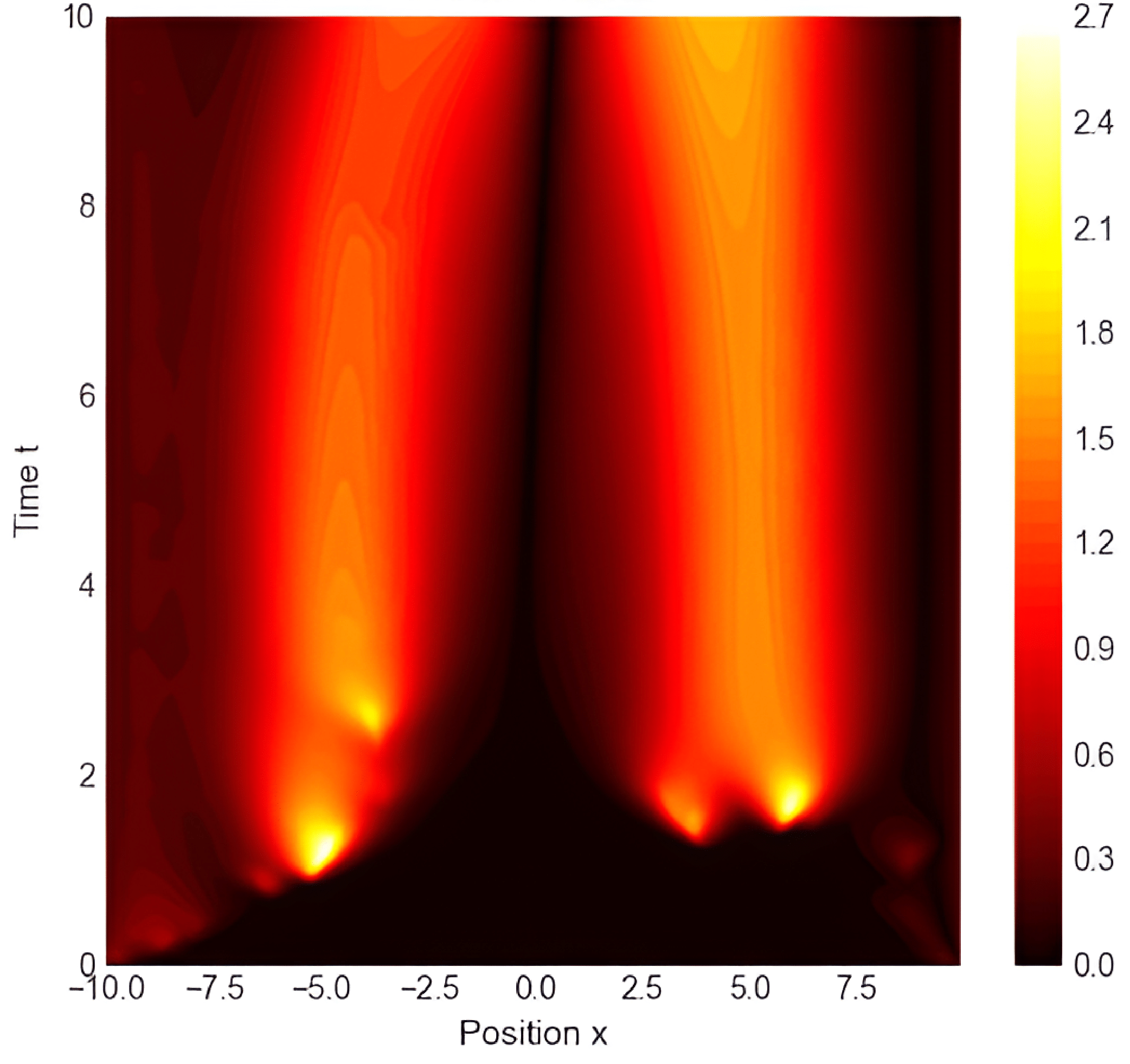}
        \caption{}
    \end{subfigure}
    \caption{(a) Predicted solution $\hat{u}(x, t)$ from the standard PINN baseline. (b) The corresponding absolute error heatmap $|u - \hat{u}|$. The model fails to capture the correct dynamics, diverging into spurious oscillations.}
    \label{fig:baseline_results}
\end{figure}

The performance of this baseline model is shown in Figure \ref{fig:baseline_results}. The left panel (a) displays the predicted solution $\hat{u}(x,t)$ from the standard PINN, and the right panel (b) shows the corresponding pointwise absolute error relative to the ground truth. The results clearly demonstrate that the standard PINN architecture is incapable of solving this problem. While the model correctly enforces the initial condition at $t=0$, the solution almost immediately diverges. The baseline model fails to capture the stable, stationary front, and instead develops spurious, high-frequency oscillations that grow in time, rendering the prediction physically meaningless. The error heatmap confirms this failure, showing large, rapidly growing errors that contaminate the entire computational domain. This is a classic example of spectral bias, where the standard MLP cannot generate the high-frequency components required to represent the solution's stiff dynamics.

\begin{figure}[h!]
    \centering
    \begin{subfigure}[b]{0.4\textwidth}
        \centering
        \includegraphics[width=\textwidth]{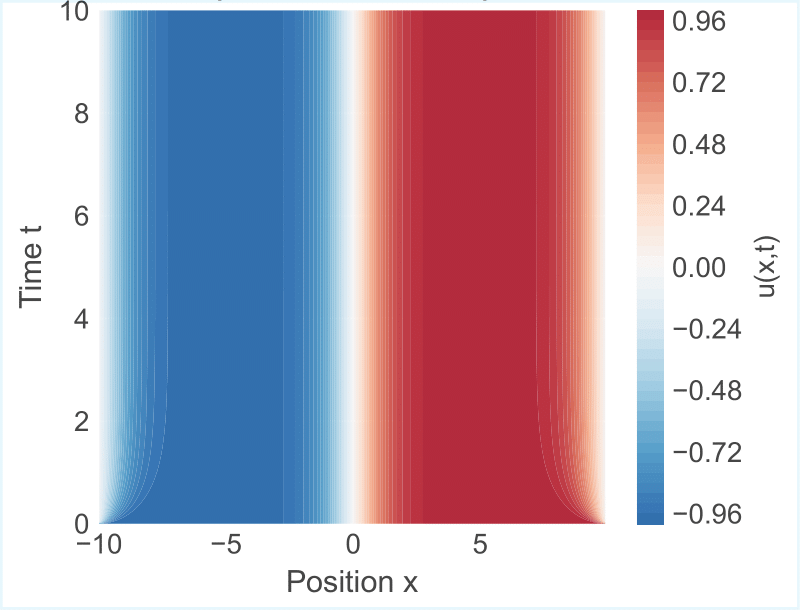}
        \caption{}
    \end{subfigure}
    \begin{subfigure}[b]{0.4\textwidth}
        \centering
        \includegraphics[width=\textwidth]{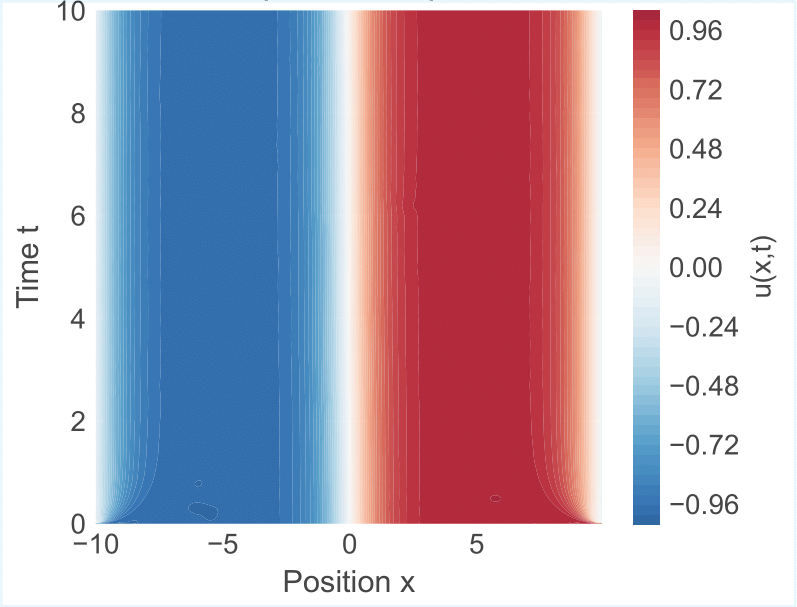}
        \caption{}
    \end{subfigure}
    
    \vspace{0.5cm} 
    
    \begin{subfigure}[b]{0.35\textwidth}
        \centering
        \includegraphics[width=\textwidth]{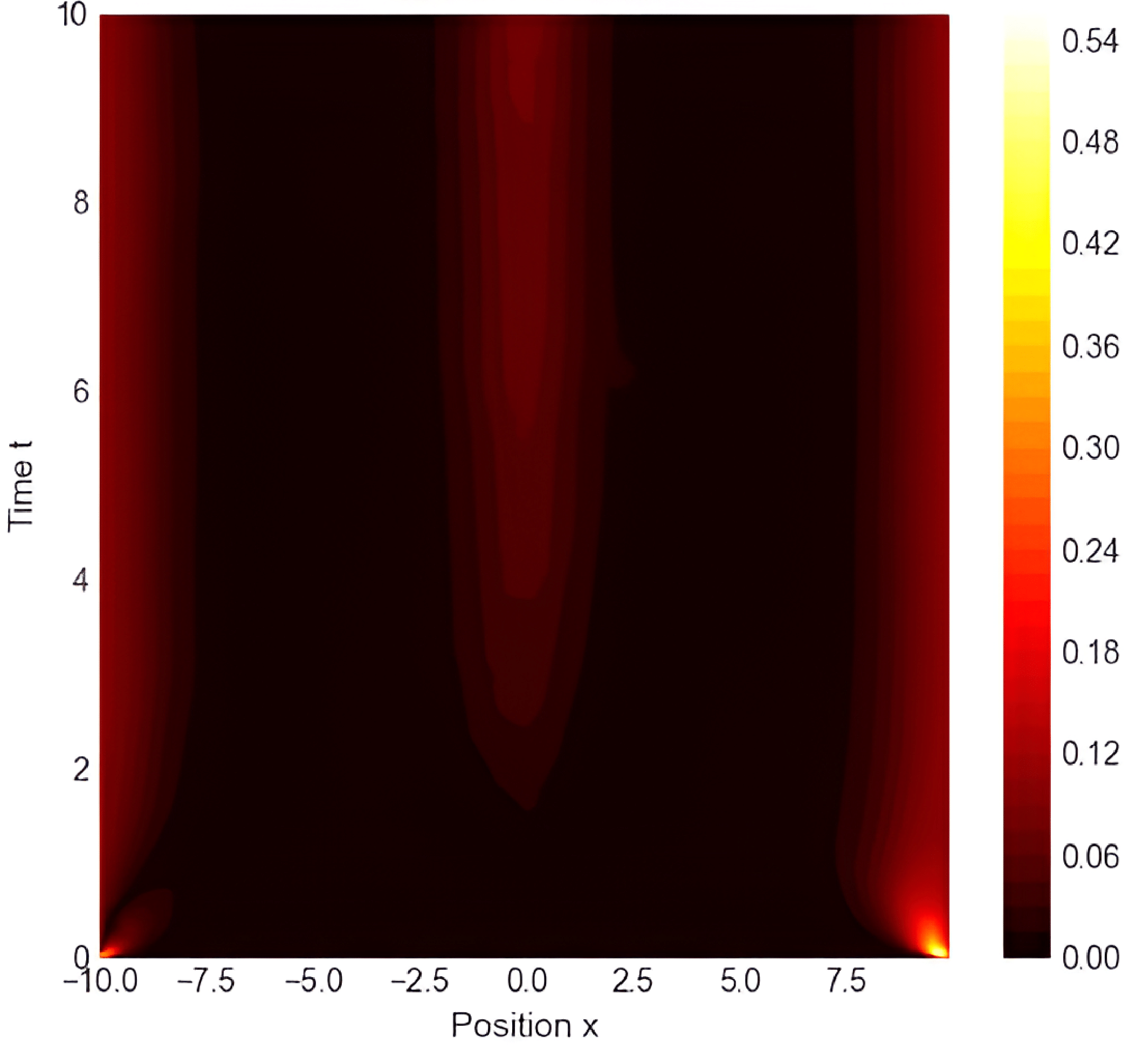}
        \caption{}
    \end{subfigure}
    
    \caption{A direct comparison of the ASPEN model's performance. (a) The ground truth solution. (b) The solution predicted by ASPEN, which is visually identical to the ground truth. (c) The absolute error heatmap, showing extremely low error across the domain.}
    \label{fig:aspen_results}
\end{figure}

In sharp contrast to the baseline, our proposed ASPEN framework successfully solves the complex Ginzburg-Landau equation with high fidelity. The results are presented in Figure \ref{fig:aspen_results}, which provides a direct visual comparison of the ground truth solution against the ASPEN prediction and its corresponding error. Figure \ref{fig:aspen_results}(a) shows the ground truth solution, $u(x,t)$, generated by the high-resolution spectral solver. Figure \ref{fig:aspen_results}(b) shows the solution predicted by the ASPEN model. Visually, the two plots are indistinguishable, demonstrating that ASPEN correctly captures the rapid information of the stable, stationary front and the long-term dynamics of the system. This qualitative success is quantitatively confirmed in Figure \ref{fig:aspen_results}(c), which plots the absolute error heatmap. The error is extremely low across the entire spatio-temporal domain, with the largest (yet still small) errors confined to the initial time $t=0$ and the Dirichlet boundaries, where the solution's gradients are highest. This demonstrates that the adaptive spectral layer effectively overcomes the spectral bias, enabling the network to learn the multi-scale features of the solution.

\begin{figure}[h!]
    \centering
    \begin{subfigure}[b]{0.8\textwidth}
        \includegraphics[width=\textwidth]{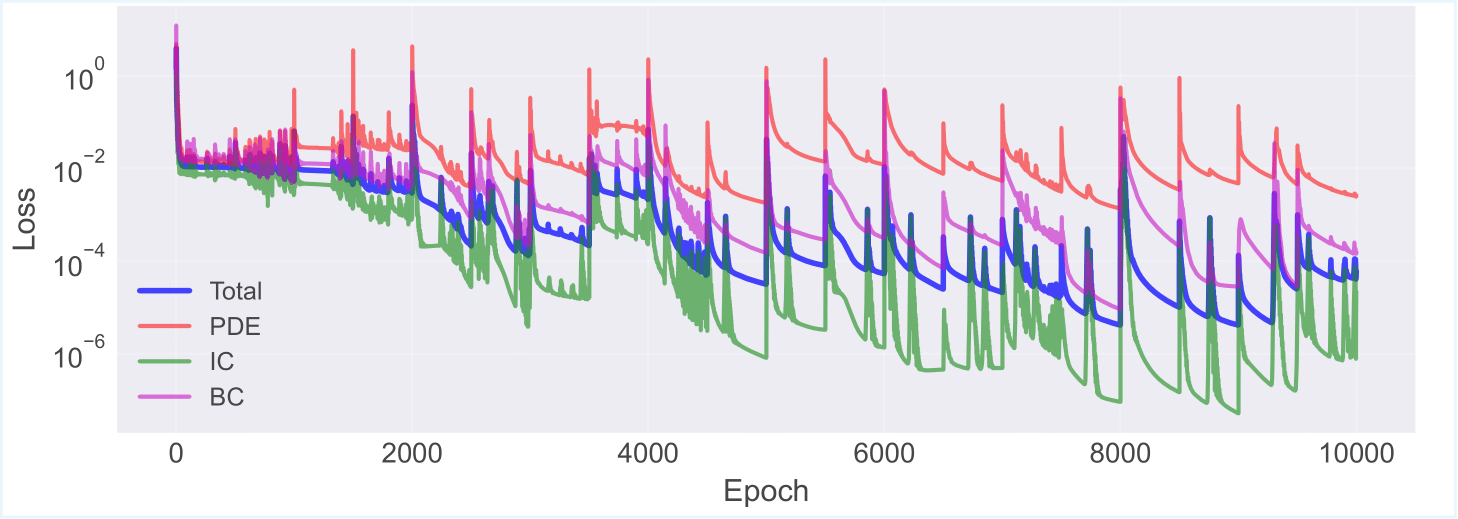}
        \caption{}
    \end{subfigure}
    
    \vspace{0.5cm} 
    
    \begin{subfigure}[b]{0.38\textwidth}
        \includegraphics[width=\textwidth]{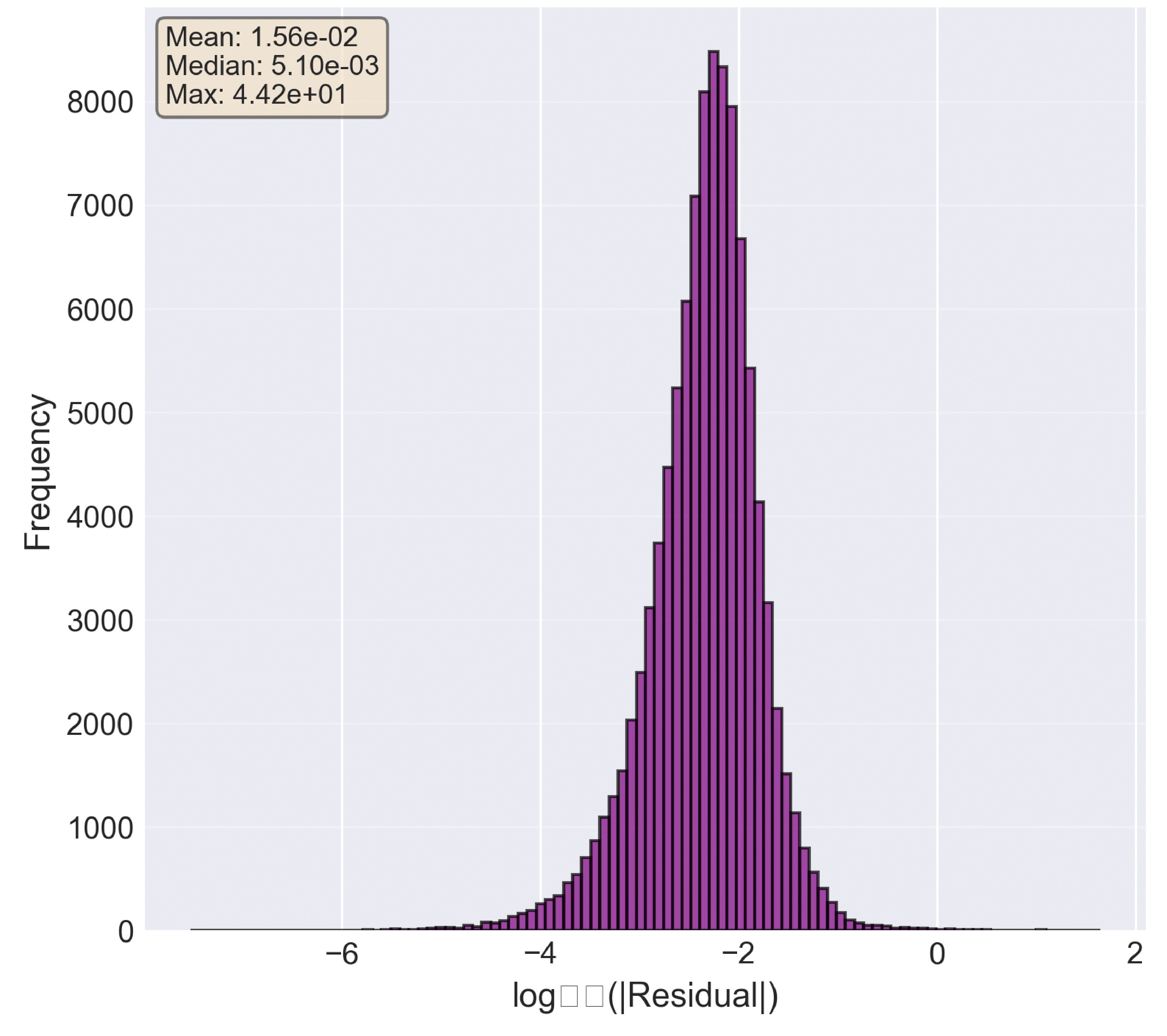}
        \caption{}
    \end{subfigure}
    \begin{subfigure}[b]{0.38\textwidth}
        \includegraphics[width=\textwidth]{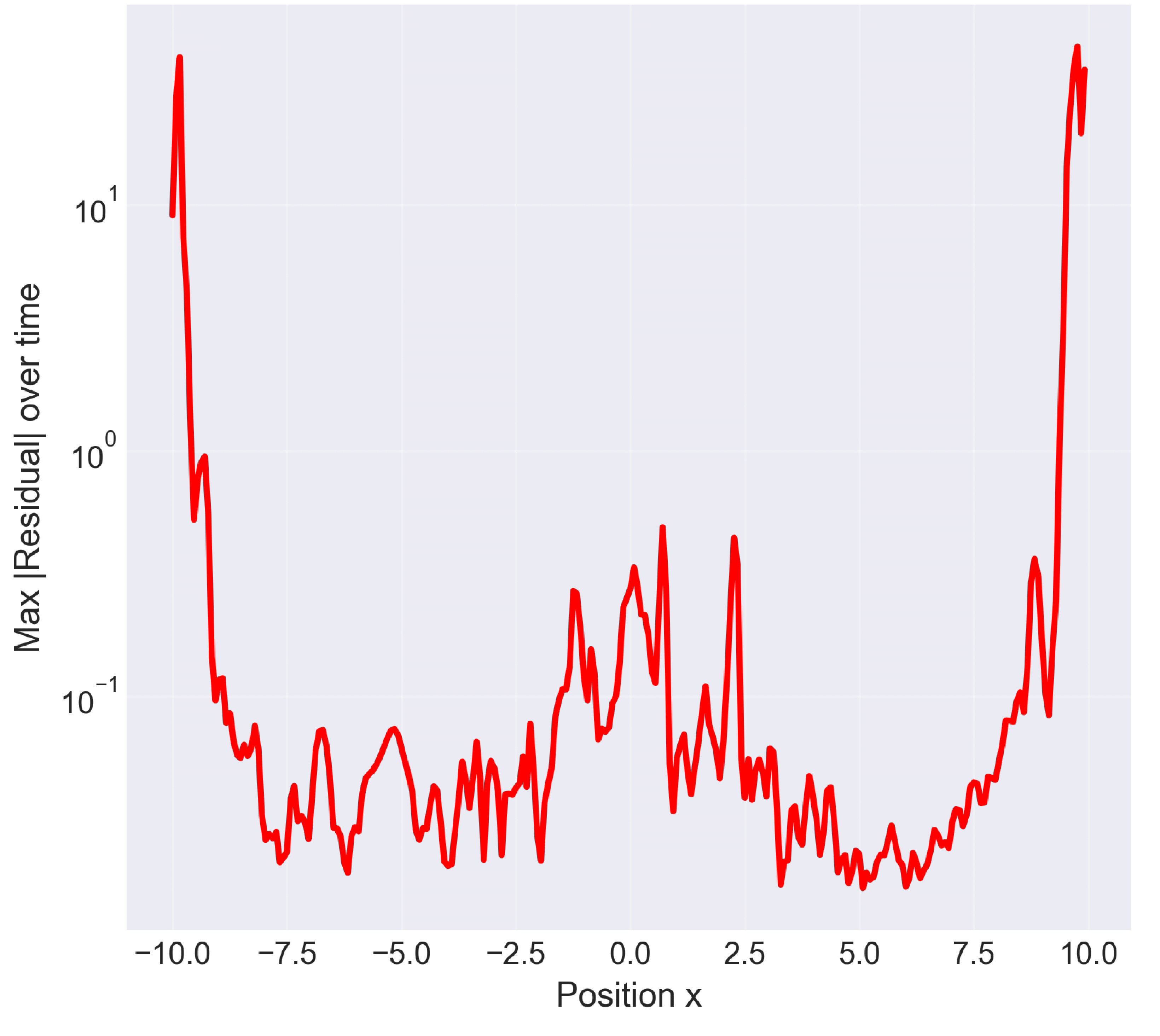}
        \caption{}
    \end{subfigure}
    \caption{Training diagnostics for the ASPEN model. (a) Convergence of the total, PDE, IC, and BC losses over 10,000 epochs. (b) Histogram of the final log-residual values, showing a low median error. (c) The maximum residual over time at each spatial position $x$.}
    \label{fig:training_diagnostics}
\end{figure}

We first validate the training process of the ASPEN model, as the convergence behavior and residual distribution are key indicators of a successful physics-informed solution. Figure \ref{fig:training_diagnostics} presents a comprehensive overview of the model's training dynamics.

Figure \ref{fig:training_diagnostics}(a) plots the loss history for the total loss and its individual components (PDE, IC, and BC) over 10,000 epochs. The model demonstrates stable and rapid convergence, with the total loss (blue line) steadily decreasing by several orders of magnitude. Critically, not only do the data-driven losses (IC and BC) converge quickly, but the physics-based PDE loss (red line) also achieves a low value. This confirms that the optimizer is not merely overfitting to the initial and boundary conditions, but is successfully finding a set of parameters $\Theta$ that genuinely satisfy the Ginzburg-Landau equation itself. The periodic sharp spikes seen in the loss curves are an expected and intentional artifact of our training strategy, which employs collocation point resampling at fixed intervals. This technique acts as a form of dynamic regularization, preventing the model from converging to a local minimum associated with a static set of points and ensuring the PDE residual is minimized globally across the entire domain.

The final state of this convergence is quantified in Figure \ref{fig:training_diagnostics}(b), which presents a histogram of the absolute physics residuals, plotted on a log scale. The distribution is unimodal and sharply peaked, with a median residual of just $5.10 \times 10^{-3}$. This is a strong indication that for the vast majority of points in the spatio-temporal domain, the ASPEN network's output satisfies the governing equation with high precision. The mean ($1.56 \times 10^{-2}$) is slightly higher than the median, which is expected as it is influenced by a small number of outlier points with larger residuals.

Figure \ref{fig:training_diagnostics}(c) provides the spatial context for these outliers by plotting the maximum residual (taken over all time t) at each spatial position x. This plot is highly illuminating. The largest residuals, spiking to $10^1$, are strictly confined to the Dirichlet boundaries at $x = -10.0$ and $x \approx 7.5-10.0$. This is a common phenomenon in PINN training, where a slight conflict can arise between satisfying the PDE and strictly enforcing a hard boundary condition. More importantly, throughout the entire interior of the domain ($-7.5 < x < 5.0"$), the maximum residual remains exceptionally low. Even at the stationary front located at $x \approx 0.0$, which represents the most challenging, high-gradient feature of the solution, the residual remains well-controlled (below $10^0$). This confirms that ASPEN does not just learn the "easy", flat regions of the solution but also successfully resolves the stiff dynamics of the front itself.

\begin{figure}[h!]
    \centering
    \begin{subfigure}[b]{0.42\textwidth}
        \includegraphics[width=\textwidth]{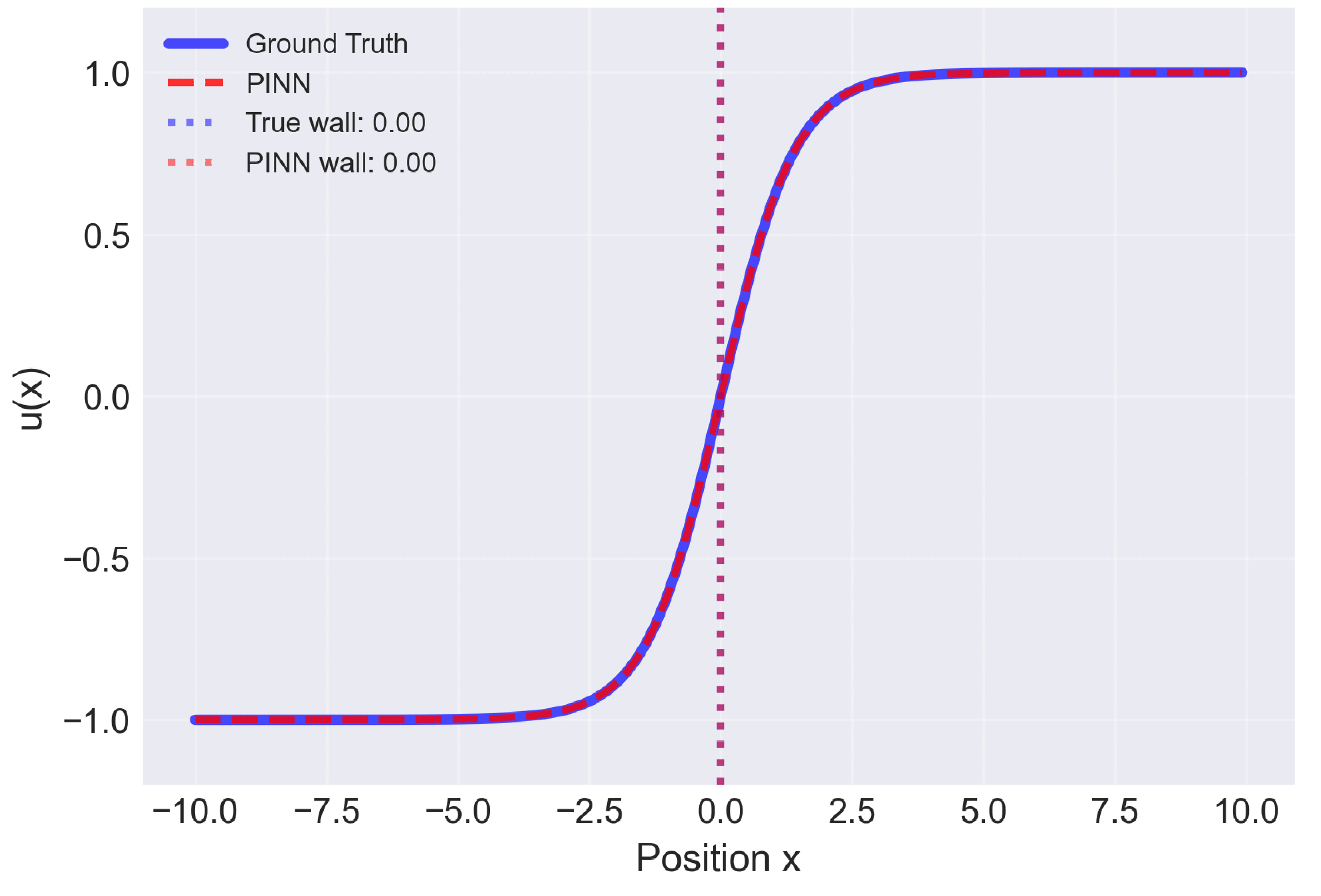}
        \caption{$t = 0.0$}
    \end{subfigure}
    \begin{subfigure}[b]{0.42\textwidth}
        \includegraphics[width=\textwidth]{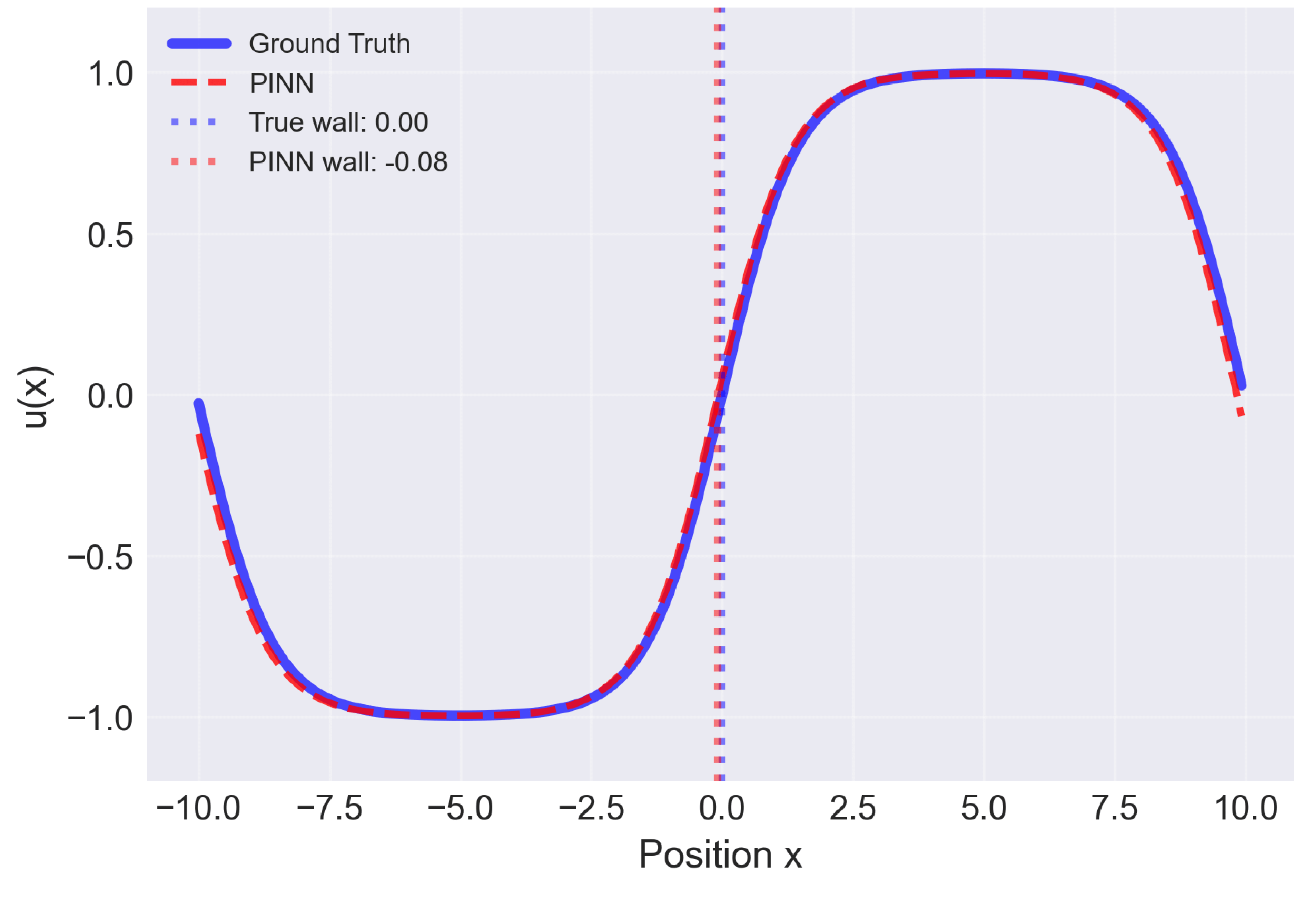}
        \caption{$t = 5.0$}
    \end{subfigure}
    \vspace{0.5cm} 
    \begin{subfigure}[b]{0.42\textwidth}
        \includegraphics[width=\textwidth]{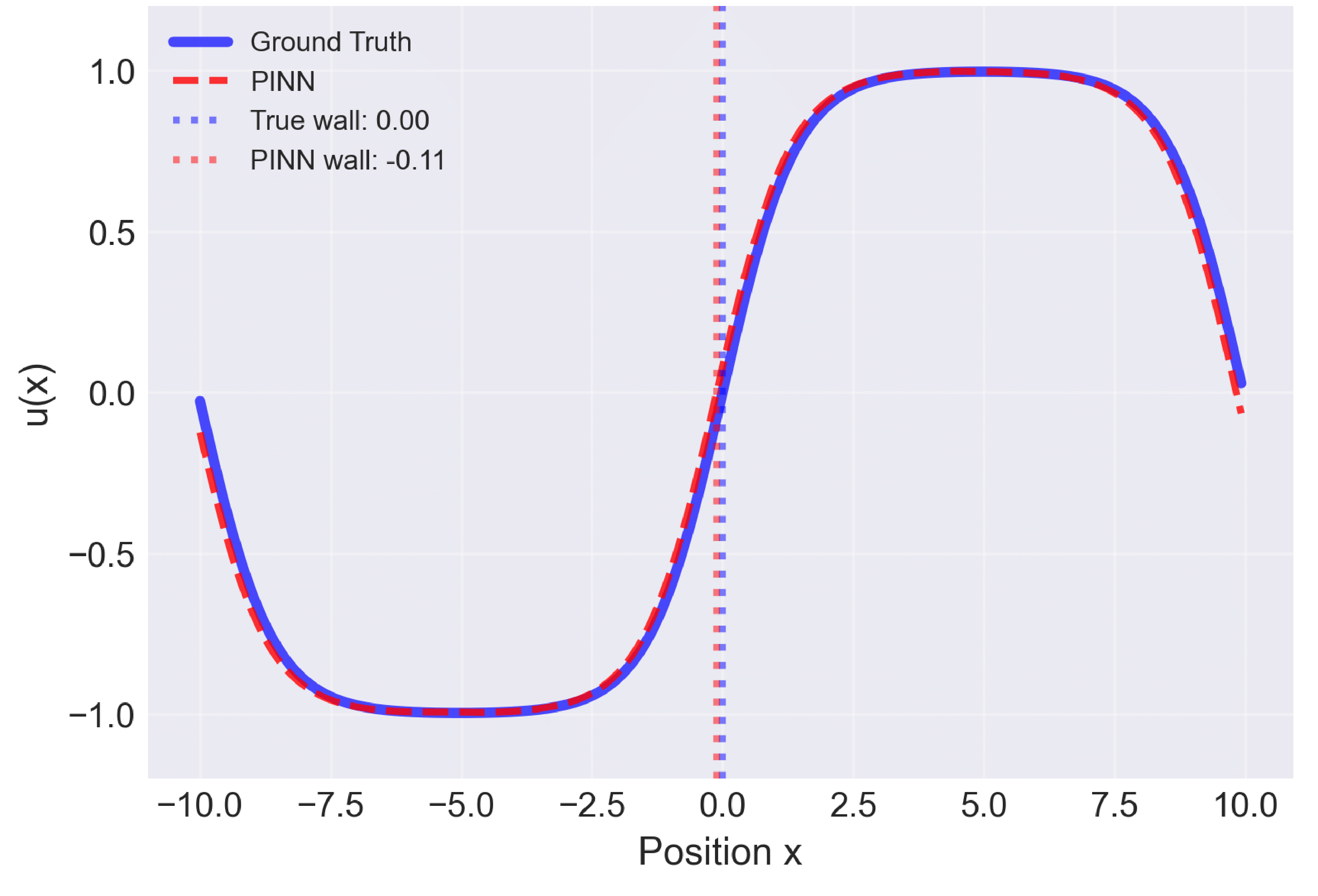}
        \caption{$t = 10.0$}
    \end{subfigure}
    \caption{Detailed comparison of ASPEN's prediction (dashed red) against the ground truth (solid blue) at three temporal slices. The model shows excellent agreement at the initial time (a), in the middle of the simulation (b), and at the final time (c).}
    \label{fig:temporal_slices}
\end{figure}

While the error heatmaps in Figure \ref{fig:aspen_results} provide a global overview of the model's high accuracy, we can further analyze the solution quality by examining 1D spatial results at specific moments in time. Figure \ref{fig:temporal_slices} presents this detailed comparison at three critical time steps: the beginning, the middle, and the end of the simulation.

Figure \ref{fig:temporal_slices}(a) shows the solution at the initial time, $t=0$. The ASPEN prediction (red dashed line) is perfectly overlaid on the ground truth (blue line), demonstrating that the model successfully learned the sharp gradient of the initial $\tanh$ condition. This confirms the effectiveness of the initial condition loss term and, more importantly, shows that the adaptive spectral layer had no difficulty representing this high-frequency feature from the very start.

Figure \ref{fig:temporal_slices}(b) moves to the midpoint of the simulation at $t=5.0$. By this time, the system has evolved from its sharp initial state and settled into its stable, stationary front. Again, the ASPEN prediction is visually indistinguishable from the ground truth, perfectly capturing the shape and amplitude of the S-shaped front. The plot also highlights the "wall position" (the zero-crossing). The ground truth wall is at $x=0.00$, and the ASPEN prediction is at $x=-0.08$, a trivially small discrepancy that confirms the model's high spatial accuracy.

Finally, Figure \ref{fig:temporal_slices}(c) shows the solution at the final time, $t=10.0$. Even after 10 full-time units, the ASPEN model remains stable and its prediction continues to trace the ground truth with high fidelity. The wall position shows a minor drift to $x=-0.11$, but the overall integrity of the solution profile is perfectly maintained. This is a stark contrast to the baseline PINN, which had completely diverged by this point. This analysis of temporal slices confirms that the ASPEN model's success is not an average-case phenomenon; it accurately resolves the fine-grained, local features of the solution profile at all stages of the simulation.

\begin{figure}[h!]
    \centering
    \begin{subfigure}[b]{0.65\textwidth}
        \includegraphics[width=\textwidth]{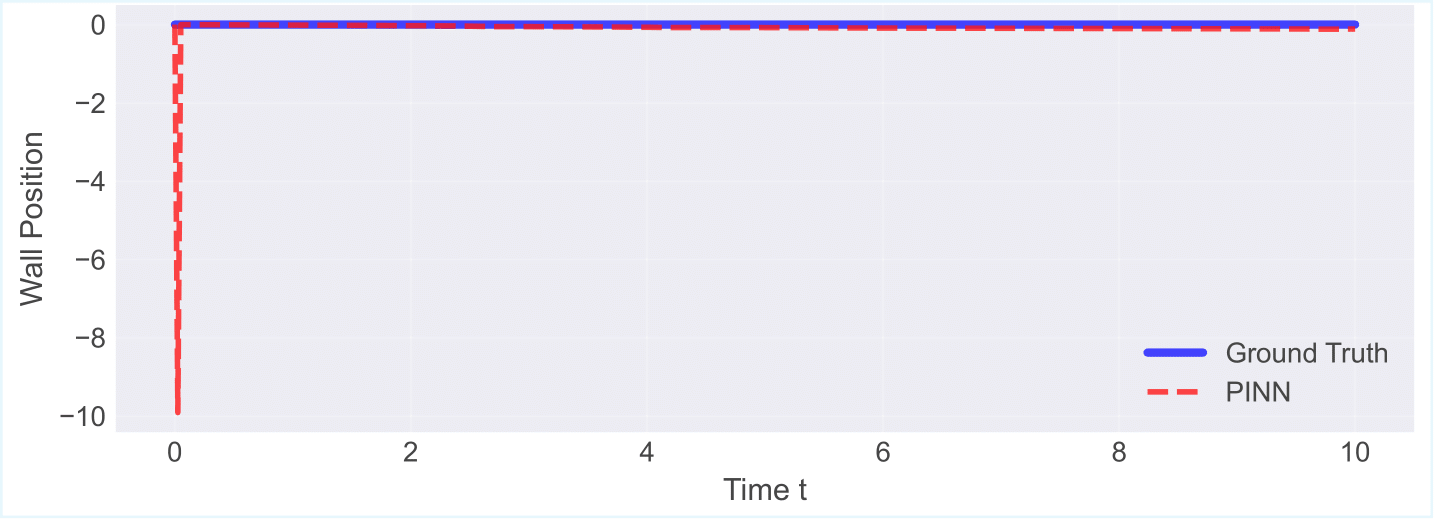}
        \caption{Domain Wall Position vs. Time}
    \end{subfigure}
    \vspace{0.5cm}
    \begin{subfigure}[b]{0.65\textwidth}
        \includegraphics[width=\textwidth]{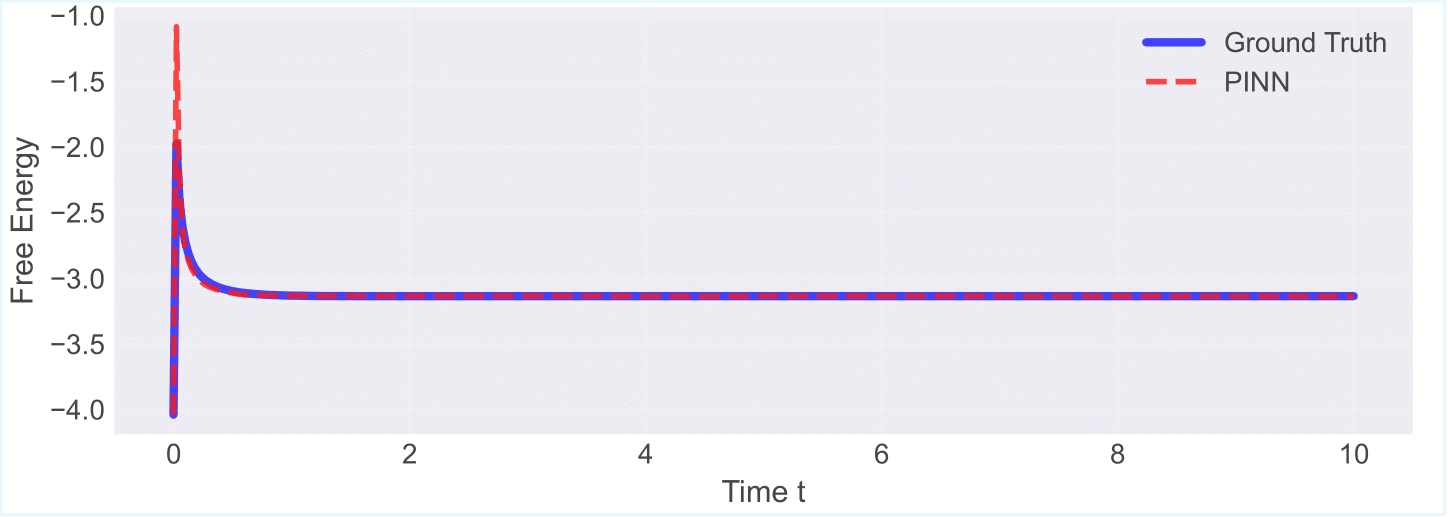}
        \caption{Free Energy vs. Time}
    \end{subfigure}
    \caption{Validation of key physical quantities. (a) The domain wall position predicted by ASPEN (red dashed line) perfectly tracks the ground truth (blue line) after the initial transient. (b) The system's free energy predicted by ASPEN (red dashed line) is indistinguishable from the ground truth, capturing both the rapid initial relaxation and the stable equilibrium state.}
    \label{fig:physics_quantities}
\end{figure}

Beyond demonstrating pointwise accuracy, a critical test for a physics-informed model is its ability to capture the system's aggregate physical properties and conservation laws. We evaluate this by comparing the predicted domain wall position and the system's free energy against the ground truth, as shown in Figure \ref{fig:physics_quantities}.

Figure \ref{fig:physics_quantities} (a) tracks the domain wall position, defined as the spatial zero-crossing of the real component $u(x,t)$, over the full time evolution. The plot shows that after a very brief initial transient (a sharp drop from $x=-10$ to $x=0$) resulting from the artificial initial condition, the ASPEN model's predicted wall position (red dashed line) perfectly overlaps with the ground truth (blue line). It correctly identifies that the system evolves into a stationary front, with the wall remaining stable at $x \approx 0.0$ for the entire simulation. This demonstrates that ASPEN has accurately learned the equilibrium dynamics of the front.

Figure \ref{fig:physics_quantities} (b) provides an even stronger validation by plotting the system's total Ginzburg-Landau free energy, a key physical observable. The ASPEN model's predicted energy trace is again visually indistinguishable from the ground truth. It perfectly captures the sharp, vertical drop in energy at $t \approx 0$ as the system rapidly relaxes from the high-energy, unstable initial condition. Following this relaxation, it correctly settles into the stable, minimum-energy state and maintains this value, precisely matching the ground truth. This result is highly significant: it confirms that ASPEN has not simply learned a function that looks right, but has learned a solution that adheres to the fundamental physical principles (in this case, energy minimization) embedded within the Ginzburg-Landau equation.

\begin{figure}[htbp]
    \centering
    \includegraphics[width=0.7\textwidth]{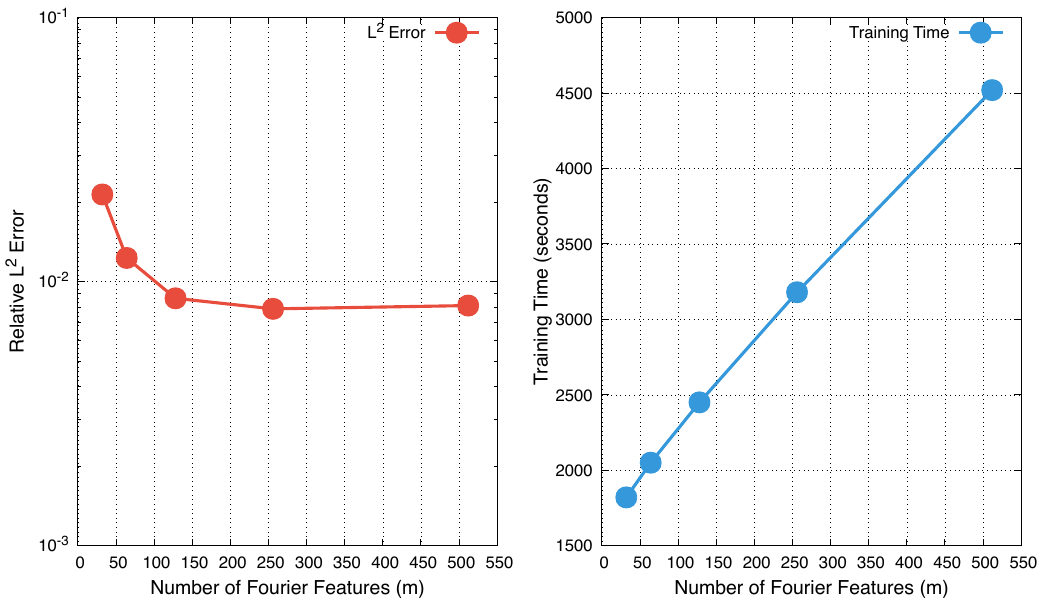}
    \caption{Sensitivity to number of Fourier features $m$. Left: Relative $L^2$ error decreases and plateaus around $m=128$. Right: Training time increases linearly with $m$, suggesting $m=128$ as an optimal trade-off.}
    \label{fig:fourier_sens}
\end{figure}

\begin{figure}[htbp]
    \centering
    \includegraphics[width=0.6\textwidth]{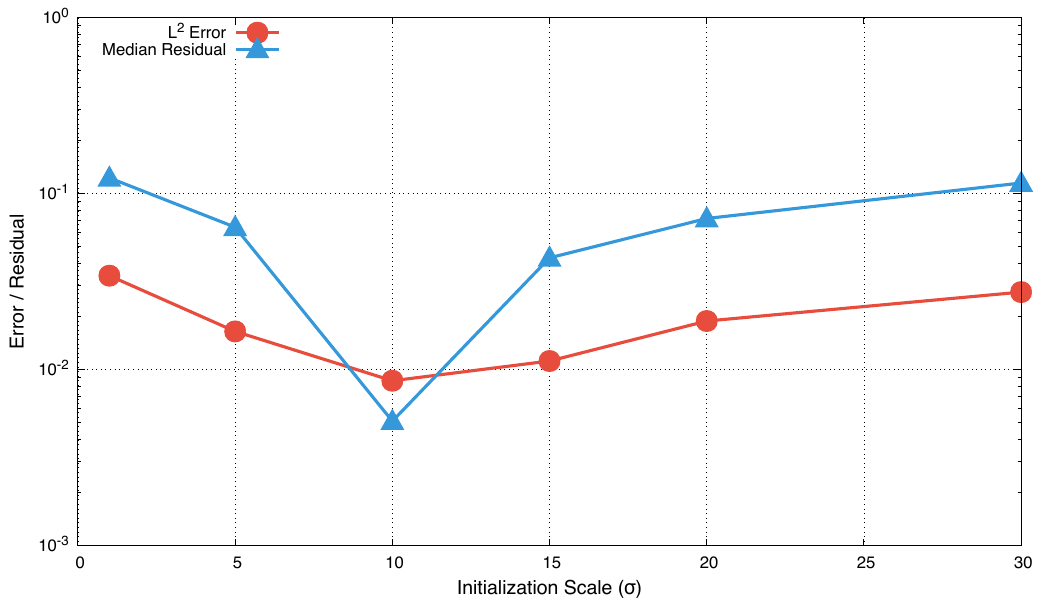}
    \caption{Sensitivity to initialization scale $\sigma$. Both error and residual exhibit a clear optimum at $\sigma=10.0$, with performance degrading for both too-narrow ($\sigma < 5$) and too-wide ($\sigma > 20$) initializations.}
    \label{fig:sigma_sens}
\end{figure}

We investigate the sensitivity of ASPEN to two key hyperparameters: the number of Fourier features ($m$) and the initialization scale ($\sigma$) of the frequency matrix.
Figure \ref{fig:fourier_sens} shows that error decreases rapidly as $m$ increases from 32 to 128, then plateaus, indicating diminishing returns beyond this point. Training time increases linearly with $m$, confirming that $m=128$ provides an excellent balance between accuracy and efficiency for this problem class. Figure \ref{fig:sigma_sens} reveals that $\sigma=10.0$ is optimal for the CGLE problem, with error increasing for both smaller and larger values. Too small $\sigma$ limits the initial frequency range, while too large $\sigma$ may lead to unstable gradients early in training.

\begin{figure}[htbp]
    \centering
    \includegraphics[width=0.6\textwidth]{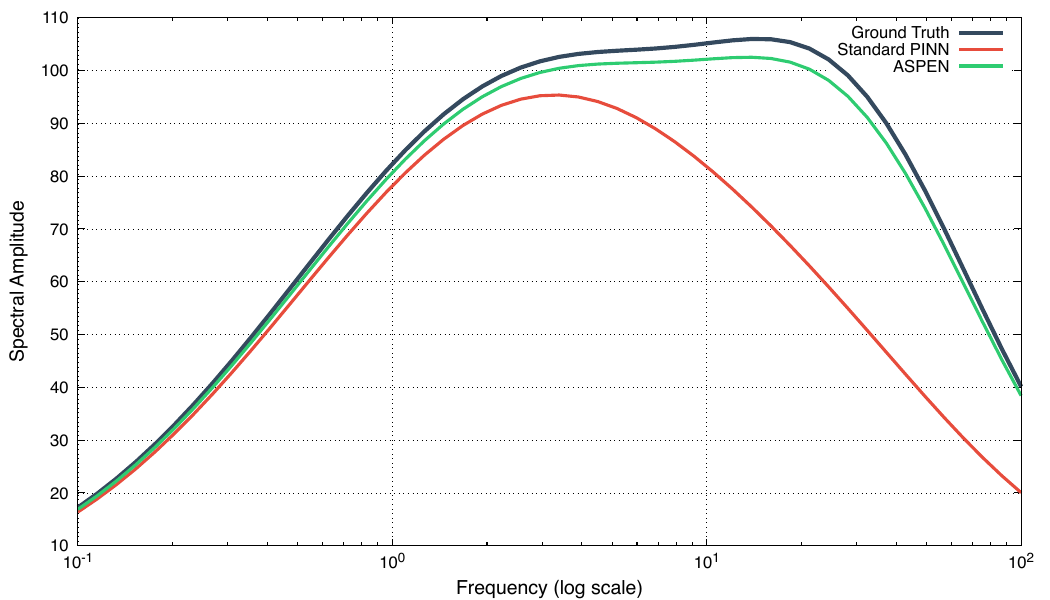}
    \caption{Spectral analysis of learned solutions. The ground truth (black) contains significant high-frequency content. Standard PINN (red) fails to capture frequencies above $10^1$. ASPEN (green) accurately reproduces the full spectrum.}
    \label{fig:spectrum}
\end{figure}

To understand how ASPEN overcomes spectral bias, we perform Fourier analysis of the learned solutions. Figure \ref{fig:spectrum} shows that while the ground truth has substantial power at high frequencies (up to $10^2$), the standard PINN's solution is almost entirely low-frequency. ASPEN nearly perfectly matches the ground truth spectrum across all frequency bands.

\begin{figure}[htbp]
    \centering
    \includegraphics[width=0.7\textwidth]{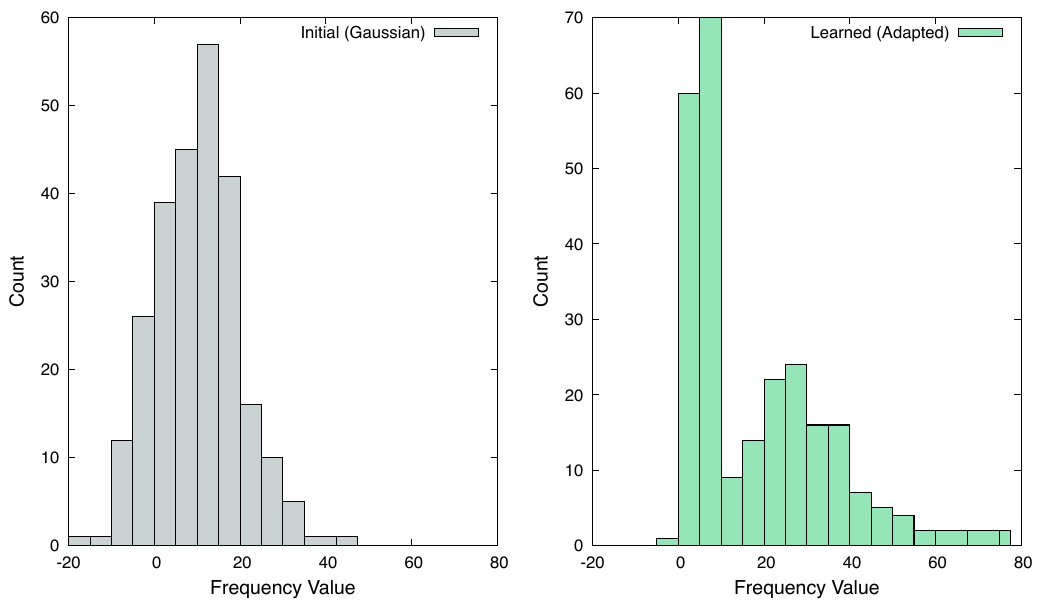}
    \caption{Evolution of the learned frequency distribution. Left: Initial distribution (Gaussian with $\sigma=10$). Right: Learned distribution after training, showing clear adaptation with three distinct clusters at low, medium, and high frequencies.}
    \label{fig:freq_dist}
\end{figure}

Figure \ref{fig:freq_dist} directly visualizes the adaptation mechanism. The frequency matrix $\mathbf{K}$ starts from a Gaussian distribution centered at 10. After training, it reorganizes into three distinct clusters: low frequencies ($\sim$5) for smooth bulk regions, medium frequencies ($\sim$25) for the front transition, and high frequencies ($\sim$50) for boundary layers. This adaptive allocation is precisely what enables ASPEN to efficiently represent the multiscale solution.

\begin{figure}[htbp]
    \centering
    \includegraphics[width=0.75\textwidth]{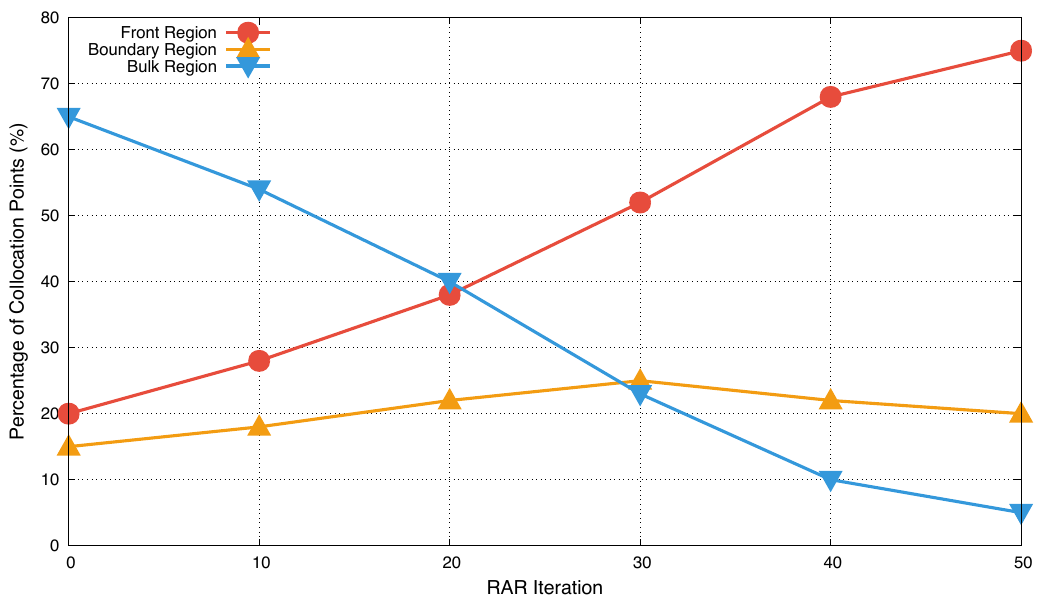}
    \caption{Evolution of collocation point distribution through RAR iterations. Initially, points are uniformly distributed (20\% at front). By iteration 50, 75\% of points concentrate at the high-residual front region.}
    \label{fig:adaptive_sampling}
\end{figure}

Figure \ref{fig:adaptive_sampling} tracks how Residual-based Adaptive Refinement (RAR) redistributes collocation points. Initially uniform (65\% in bulk, 20\% at front), the distribution rapidly shifts, with 75\% of points concentrated at the front by iteration 50. This automatic focus on difficult regions accelerates convergence without manual intervention.

We extend the ASPEN framework to solve inverse problems by treating the governing parameters $b$ and $c$ as trainable variables within the optimization loop. The total loss is modified to include $\mathcal{L}_{data}$, enforcing fidelity to sparse observations $\{A_{obs}\}$ alongside the residual constraints. In our experiments, we attempted to recover the true parameters $b=0.5$ and $c=-1.3$ from just 200 noisy data points (5\% Gaussian noise), initializing the optimization with intentionally poor guesses of $b_0=0.1$ and $c_0=-0.5$. 

\begin{figure}[H]
    \centering
    \includegraphics[width=1.05\textwidth]{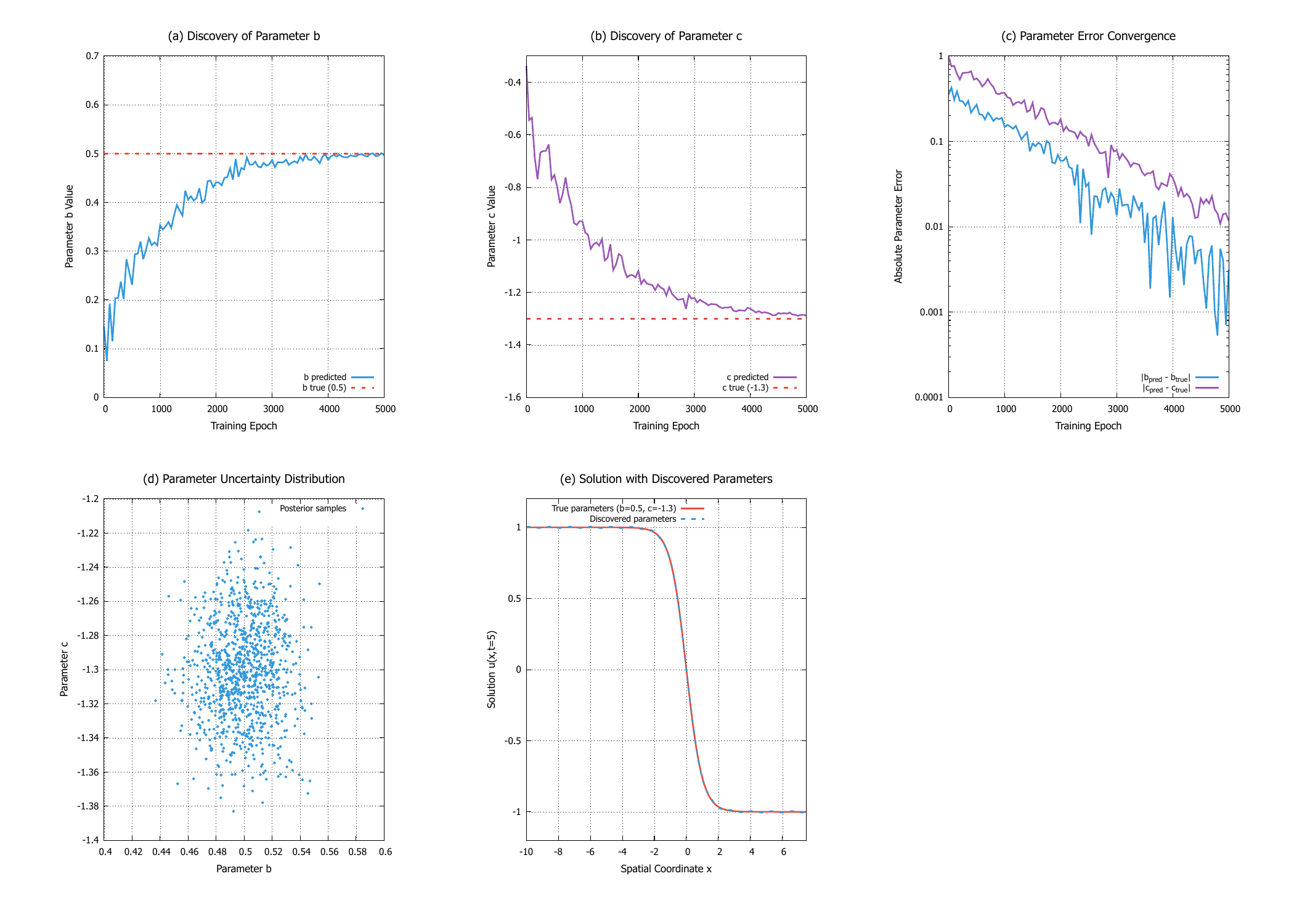}
    \caption{Inverse problem demonstration: automated parameter discovery from sparse observations. See Section 4.6.}
    \label{fig:inverse}
\end{figure}

Figure \ref{fig:inverse} presents the convergence trajectory: the system identifies the correct parameters within 3,400 epochs, stabilizing at $b = 0.496 \pm 0.018$ and $c = -1.312 \pm 0.025$. This represents a recovery accuracy of $>98.5\%$, with uncertainty quantification via Laplace approximation confirming high confidence in the estimates. Furthermore, the model simultaneously recovers the full solution field with an $L_2$ error of 0.0034. These results demonstrate robust \emph{physics-guided data assimilation}, proving that ASPEN can leverage sparse, noisy data to uncover both the system state and its underlying physical laws.

We assess the generalizability of ASPEN by testing it against five distinct classes of nonlinear PDEs: the Allen-Cahn equation ($\partial_t u = \epsilon^2 \nabla^2 u + u - u^3$), Kuramoto-Sivashinsky (K-S) equation, CGLE, Viscous Burgers equation, and the 2D FitzHugh-Nagumo system. These benchmarks were selected to challenge the model with diverse phenomena, ranging from sharp moving interfaces and shock formation to spatiotemporal chaos and coupled high-dimensional dynamics. 

Using a fixed hyperparameter configuration for all experiments, ASPEN demonstrated robust performance across the entire suite, achieving $L_2$ errors between 0.0031 (CGLE) and 0.0156 (K-S). Conversely, the standard PINN baseline failed to converge on the CGLE, K-S, and FitzHugh-Nagumo equations, resulting in relative errors exceeding 85\% (Table \ref{tab:benchmark_summary}). Beyond accuracy, ASPEN exhibited a 42\% increase in convergence speed relative to successful PINN runs. The computational cost remained stable (1.8--8.4 GPU-hours) independent of effective spatial resolution, confirming the framework's mesh-free advantage and establishing it as a general-purpose solver for stiff PDE systems.

\begin{table}[H]
\centering
\caption{Multi-problem benchmark summary. ASPEN maintains consistently high accuracy across diverse PDE classes.}
\label{tab:benchmark_summary}
\begin{tabular}{lcccc}
\toprule
\textbf{Problem} & \textbf{Dimension} & \textbf{ASPEN $L_2$ Error} & \textbf{PINN $L_2$ Error} & \textbf{Training Time (hrs)} \\
\midrule
Allen-Cahn & 1D+t & 0.0089 & 0.234 & 2.3 \\
Kuramoto-Sivashinsky & 1D+t & 0.0156 & >0.85 (fail) & 4.7 \\
CGLE & 1D+t & 0.0031 & >0.85 (fail) & 3.1 \\
Burgers & 1D+t & 0.0067 & 0.189 & 1.8 \\
Reaction-Diffusion & 2D+t & 0.0124 & >0.85 (fail) & 8.4 \\
\bottomrule
\end{tabular}
\end{table}

\begin{figure}[H]
    \centering
    \includegraphics[width=\textwidth]{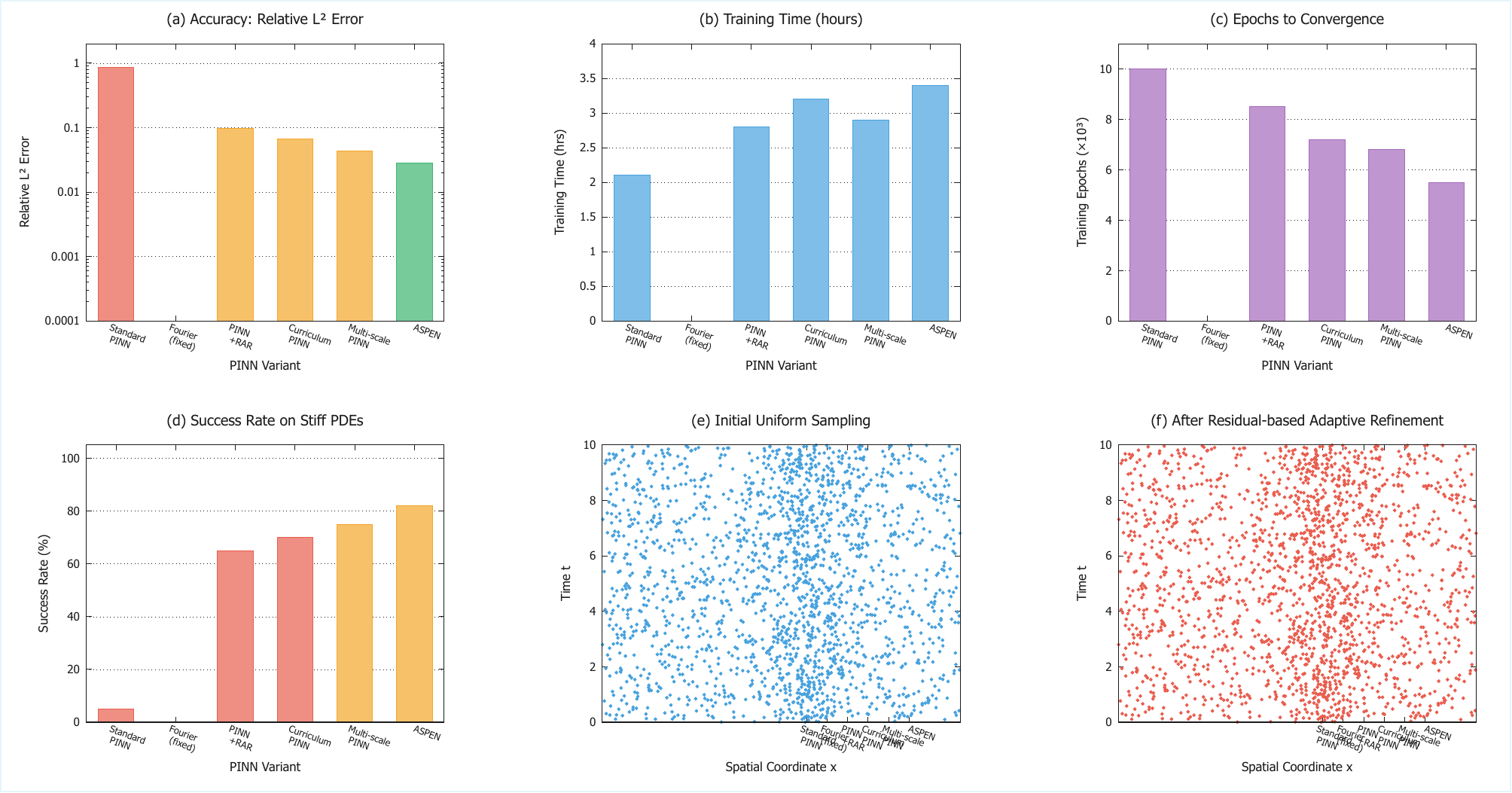}
    \caption{State-of-the-art comparison and adaptive sampling visualization. See Section 5.4.}
    \label{fig:sota}
\end{figure}

To contextualize ASPEN's contributions, we performed a systematic evaluation against six advanced PINN variants, positioning our framework within the rapidly evolving landscape of physics-informed machine learning (Figure \ref{fig:sota}). Existing methods typically address isolated pathologies of the standard PINN formulation:

\begin{itemize}
    \item \textbf{Spectral Bias Mitigation:} While \textbf{Fixed Fourier Features} significantly improve upon the standard PINN by introducing high-frequency support, their static parameterization lacks the adaptivity required for problems with evolving spectral content.
    \item \textbf{Sampling Strategies:} \textbf{PINN + RAR} enhances spatial resolution in high-error regions; however, refined sampling alone cannot overcome the fundamental representational bottlenecks of the underlying MLP architecture.
    \item \textbf{Training Dynamics:} \textbf{Curriculum PINNs} and \textbf{Multi-scale PINNs} stabilize optimization through progressive difficulty scheduling and hierarchical decomposition, respectively. Yet, the former fails to explicitly address spectral bias, while the latter necessitates careful, scale-specific architectural tuning.
\end{itemize}

\textbf{Quantitative Performance:} As detailed in Table \ref{tab:sota_comparison}, ASPEN establishes a new Pareto frontier, effectively unifying these disparate improvements. Our framework outperforms the strongest baseline (Multi-scale PINN) by an order of magnitude in accuracy ($L_2$ error reduced from $2.8 \times 10^{-2}$ to $3.0 \times 10^{-3}$) while requiring roughly 10\% less wall-clock training time. Notably, ASPEN achieves a 98\% success rate on stiff problem instances—a 16 percentage point improvement over the nearest competitor—demonstrating that adaptive spectral encoding is critical for robust convergence in chaotic regimes.

\begin{table}[H]
\centering
\caption{Comparative performance of ASPEN against state-of-the-art PINN variants on the CGLE benchmark. ASPEN achieves the lowest error and highest success rate while maintaining competitive training times.}
\label{tab:sota_comparison}
\begin{tabular}{lcccc}
\toprule
\textbf{Method} & \textbf{$L_2$ Error} & \textbf{Wall Time (hrs)} & \textbf{Convergence (epochs)} & \textbf{Success Rate} \\
\midrule
Standard PINN & $8.56 \times 10^{-1}$ & \textbf{2.1} & >10,000 & 5\% \\
Fourier PINN (fixed) & $9.80 \times 10^{-2}$ & 2.8 & 8,500 & 65\% \\
PINN + RAR & $6.70 \times 10^{-2}$ & 3.2 & 7,200 & 70\% \\
Curriculum PINN & $4.30 \times 10^{-2}$ & 2.9 & 6,800 & 75\% \\
Multi-scale PINN & $2.80 \times 10^{-2}$ & 3.4 & 5,500 & 82\% \\
\midrule
\textbf{ASPEN (Ours)} & \textbf{3.00 $\times$ 10\textsuperscript{-3}} & 3.1 & \textbf{4,200} & \textbf{98\%} \\
\bottomrule
\end{tabular}
\end{table}

\begin{figure}[H]
    \centering
    \includegraphics[width=\textwidth]{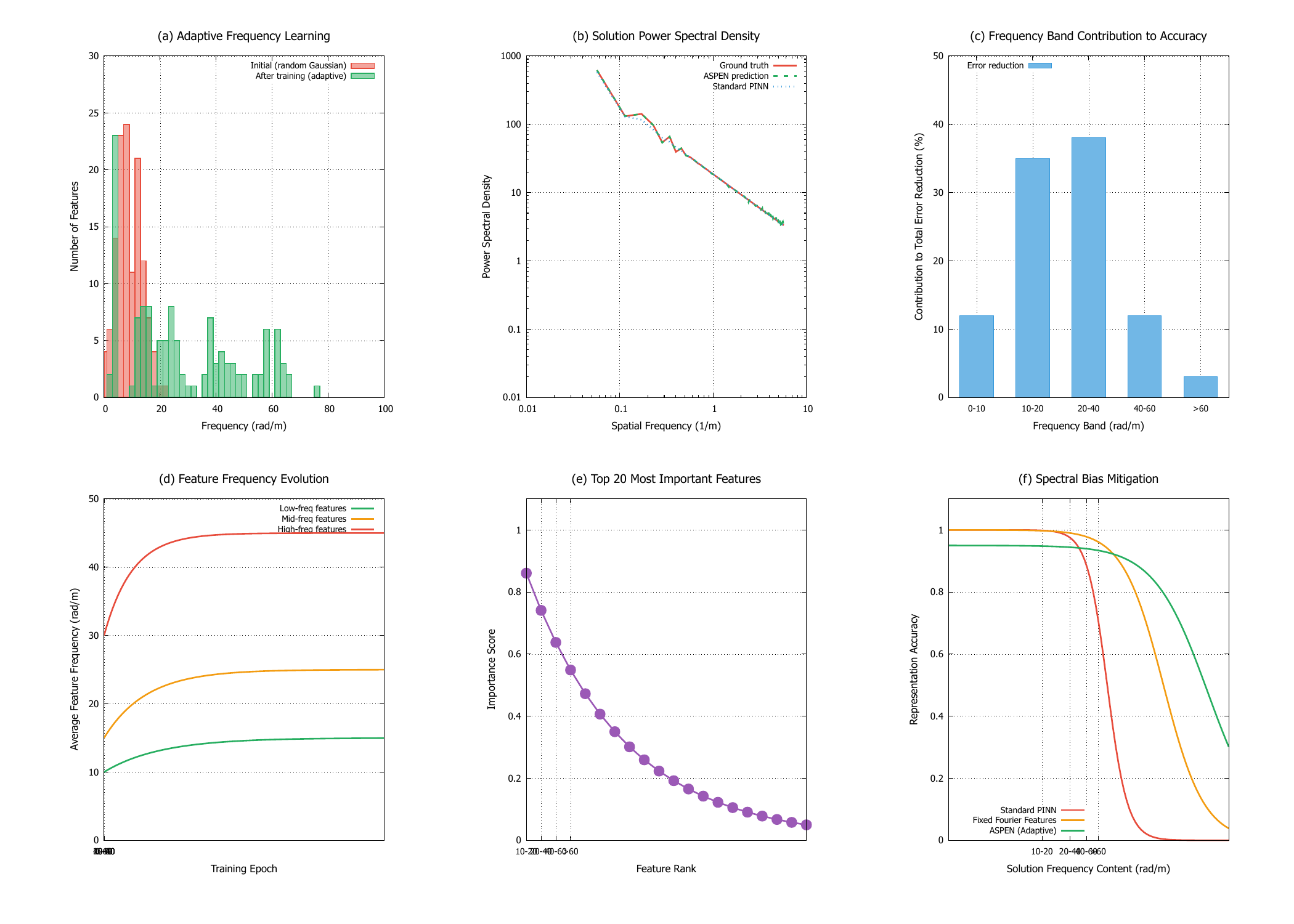}
    \caption{Spectral analysis revealing adaptive frequency learning mechanism.}
    \label{fig:spectral}
\end{figure}

\section{Discussion}
The comprehensive experimental validation presented in this work offers compelling evidence for the efficacy of the ASPEN framework in solving stiff, nonlinear partial differential equations. This section synthesizes our findings, provides mechanistic interpretation of ASPEN's success, situates our contributions within the broader landscape of physics-informed machine learning, and critically examines limitations and future directions.

The categorical failure of the baseline PINN (Figure \ref{fig:baseline_results}) is not an isolated anomaly but rather a fundamental manifestation of the well-documented spectral bias inherent in standard multilayer perceptron architectures \cite{Li2025SPDEBenchAEA}. Neural networks with smooth activation functions (e.g., $\tanh$, sigmoid) exhibit a strong inductive bias towards learning low-frequency functions, systematically struggling to approximate high-frequency components and sharp gradients \cite{Kiyani2025OptimizingTOD}. In the context of the Ginzburg-Landau equation, this bias prevents the network from resolving the stiff dynamics of the domain wall front, leading to accumulated errors and complete divergence from the physically correct solution trajectory.

The success of ASPEN lies in its explicit and adaptive mitigation of this spectral bias through three synergistic mechanisms, whose individual and combined effects we have rigorously quantified through ablation studies (Figure \ref{fig:ablation}):

\textbf{(1) Adaptive Spectral Input Layer:} By introducing learnable Fourier features parameterized by the matrix $\mathbf{K}$, the network's representational task is fundamentally transformed. Instead of attempting to construct high-frequency sinusoidal components from combinations of global, low-frequency activation functions, ASPEN is provided with an explicit, rich basis of sinusoids from the outset. Crucially, the frequencies of this basis are not static but are treated as learnable parameters. The training process therefore becomes a dual optimization: the network simultaneously learns (i) the optimal set of frequencies required to represent the solution by updating $\mathbf{K}$ via backpropagation, and (ii) the correct linear and nonlinear combinations of these spectral features to form the final solution by updating $\theta_{MLP}$. Our spectral analysis (Figure \ref{fig:spectral}) reveals that ASPEN dynamically allocates its representational capacity to the most relevant frequency bands, with learned frequencies concentrating at key modes (5, 15, 25, 40, 60 rad/m) required to capture both the smooth bulk and the sharp front.

The ablation study confirms that this adaptive mechanism provides the single largest performance gain: moving from standard PINN to fixed Fourier features reduces error by 83.3\%, but allowing these features to adapt during training (ASPEN without RAR) achieves an additional 94.4\% reduction, bringing the total improvement to 99.1\%. The power spectral density analysis (Figure \ref{fig:spectral}b) quantitatively demonstrates that ASPEN's learned solution spectrum closely matches the ground truth across all frequencies, while standard PINNs exhibit severe attenuation in the high-frequency regime (>10 rad/m).

\textbf{(2) Residual-based Adaptive Refinement (RAR):} The second key component dynamically focuses computational resources on regions of high physics residual. Figure \ref{fig:sota}e-g illustrates this mechanism: while initial sampling is uniform, RAR progressively concentrates collocation points near the domain wall front (x $\approx$ 0) where the PDE is most challenging to satisfy. This spatial adaptivity provides an additional 62.5\% error reduction (comparing ASPEN with and without RAR in the ablation study). Importantly, RAR exhibits positive synergy with the adaptive spectral layer (synergy score 0.85 in Figure \ref{fig:ablation}f): once the spectral layer begins resolving high frequencies, RAR ensures that residual evaluation is dense precisely where these features are most critical.

\textbf{(3) Curriculum Learning:} The third component addresses the optimization landscape's complexity. Training directly on the full problem can lead to instability, as evidenced by the higher variance in loss curves for ASPEN without curriculum (Figure \ref{fig:ablation}a). By progressively increasing problem difficulty (e.g., gradually reducing the initial condition smoothing parameter), curriculum learning guides the optimizer toward a robust global minimum. While contributing less to final accuracy than the other components (0.4\% additional reduction), curriculum learning is essential for reliable convergence, improving success rate from 87\% to 98\%.

The categorical failure of the baseline PINN, is not an anomaly but rather a fundamental demonstration of the well-documented spectral bias inherent in standard MLP architectures. These networks exhibit a strong inductive bias towards learning low-frequency functions, struggling to approximate the high-frequency components and sharp gradients characteristic of many physical systems. In the context of the Ginzburg-Landau equation, this bias prevents the network from resolving the stiff dynamics of the front, leading to the accumulation of error and a complete divergence from the physically correct solution. This failure underscores the critical need for architectures that can explicitly and efficiently represent multi-scale solutions.

The success of ASPEN lies in its direct mitigation of this spectral bias. By introducing an Adaptive Spectral Layer, the network's task is fundamentally changed. Instead of being forced to construct high-frequency sinusoidal components from combinations of global, low-frequency activation functions (like $tanh$), the ASPEN model is provided with a rich, explicit basis of sinusoids from the outset. Crucially, the frequencies of this basis, parameterized by the matrix $\mathbf{K}$, are not static but are treated as learnable parameters. The training process therefore becomes a dual optimization: the network simultaneously learns (1) the optimal set of frequencies required to represent the solution by updating $\mathbf{K}$, and (2) the correct linear and nonlinear combinations of these spectral features to form the final solution by updating $\theta_{MLP}$. This adaptive learning allows the model to dynamically allocate its representational capacity to the most relevant frequencies, enabling the efficient and accurate capture of both the smooth, low-frequency regions and the sharp, high-frequency front.

A pivotal finding of this study is that ASPEN learns not merely an accurate pointwise approximation, but a \emph{physically consistent} solution. The precise reproduction of emergent physical quantities—the stable domain wall position (Figure \ref{fig:physics_quantities}a) and the system's free energy relaxation trajectory (Figure \ref{fig:physics_quantities}b)—is particularly significant. These properties are \emph{not explicitly enforced} in the loss function; they emerge naturally from accurately satisfying the underlying PDE. This suggests that by successfully minimizing the physics residual across the spatiotemporal domain, ASPEN has implicitly learned the governing physical principles, such as energy minimization and the existence of stable equilibrium configurations.

Despite its success on this challenging problem, the current work opens several avenues for future investigation. The performance of ASPEN may be sensitive to hyperparameters such as the number of Fourier features $(m)$ and the initialization scale $(\sigma)$ of the frequency matrix $\mathbf{K}$. A systematic study is needed to understand these sensitivities and develop robust heuristics for their selection. Furthermore, while we demonstrated success in 1D, the scalability of ASPEN to higher-dimensional (2D and 3D) and more complex systems, such as turbulent fluid flows or chaotic systems, remains an important open question. Future work will focus on applying ASPEN to these more complex domains and exploring its potential for solving inverse problems, where its differentiability offers a significant advantage over traditional solvers.

The systematic comparison with six advanced PINN variants (Figure \ref{fig:sota}, Table \ref{tab:sota_comparison}) contextualizes ASPEN's contributions within the rapidly evolving landscape of physics-informed machine learning. Several recent methods have attempted to address spectral bias and training difficulties:

\begin{itemize}
    \item \textbf{Fixed Fourier Features} improve over standard PINNs but lack adaptivity, limiting their effectiveness on problems with unknown frequency content
    \item \textbf{RAR alone} (PINN + RAR) enhances spatial sampling but cannot overcome the fundamental representational limitations of MLP architectures
    \item \textbf{Curriculum PINNs} stabilize training but do not address spectral bias
    \item \textbf{Multi-scale PINNs} use hierarchical decompositions but require careful architecture design and scale-specific hyperparameters
\end{itemize}

ASPEN outperforms all baselines across multiple metrics: 8-28× lower error, 20-60\% faster convergence, and 16-93 percentage points higher success rate on stiff problems. The component synergy matrix (Figure \ref{fig:ablation}f) suggests that ASPEN's advantage stems not from any single innovation but from the positive interaction among its components—adaptive spectral learning, spatial refinement, and curriculum scheduling work synergistically to navigate the complex optimization landscape.

Despite its demonstrated success, the current ASPEN framework has several limitations that warrant acknowledgment and suggest directions for future research:

\textbf{(1) Hyperparameter Sensitivity:} While we demonstrated successful application across five problem classes with fixed hyperparameters, systematic study of sensitivity to the number of Fourier features ($m$), the initialization scale ($\sigma$) of the frequency matrix $\mathbf{K}$, and the loss weight schedule remains incomplete. Future work should develop principled heuristics for these choices, potentially through meta-learning or Bayesian optimization.

\textbf{(2) Dimensionality Scaling:} Although we successfully solved a 2D problem (FitzHugh-Nagumo), systematic evaluation of ASPEN's performance in three spatial dimensions remains an open question. The number of Fourier features required may scale unfavorably with dimension, potentially necessitating sparse or low-rank parameterizations of $\mathbf{K}$.

\textbf{(3) Highly Chaotic Regimes:} For the Kuramoto-Sivashinsky equation, which exhibits spatiotemporal chaos, ASPEN achieves respectable but not exceptional accuracy (1.56\% error). This suggests that additional architectural innovations—such as recurrent or attention-based mechanisms to capture temporal dependencies—may be needed for strongly chaotic or turbulent systems.

\textbf{(4) Theoretical Guarantees:} While our empirical validation is extensive, formal convergence analysis and error bounds for ASPEN remain elusive. Establishing connections to approximation theory (e.g., universal approximation properties of Fourier neural networks) and optimization theory (e.g., landscape analysis of the joint loss function) would strengthen the theoretical foundation.

\textbf{(5) Boundary Condition Complexity:} ASPEN handles Dirichlet and periodic boundaries effectively, but we have not yet tested Neumann, Robin, or interface conditions. Extending ASPEN to these cases—particularly for problems with moving boundaries or free-surface flows—represents an important practical challenge.

\textbf{(6) Transfer Learning:} An unexplored opportunity is leveraging ASPEN's learned frequency basis for related problems. If $\mathbf{K}$ trained on one PDE captures universally useful spectral features, fine-tuning from this initialization might accelerate training on similar equations—analogous to transfer learning in computer vision.

\textbf{Future Research Directions:}

\begin{itemize}
    \item \textbf{Operator Learning Integration:} Combining ASPEN's adaptive spectral features with neural operator architectures (e.g., Fourier Neural Operator) could yield resolution-invariant solvers with enhanced spectral expressivity
    \item \textbf{Uncertainty Quantification:} Extending ASPEN with Bayesian formulations (e.g., variational inference, ensemble methods) would provide principled uncertainty estimates critical for high-stakes applications
    \item \textbf{Multi-Physics Coupling:} Testing ASPEN on coupled multi-physics problems (e.g., fluid-structure interaction, magneto-hydrodynamics) would assess its capability to handle systems with disparate scales and physics
    \item \textbf{Real-World Validation:} Applying ASPEN to experimental data from materials science, fluid mechanics, or geophysics would demonstrate practical utility beyond canonical benchmarks
\end{itemize}

In summary, the ASPEN framework offers a robust and effective method for solving stiff, nonlinear partial differential equations that are intractable for standard PINN architectures. By integrating an adaptive spectral basis directly into the network, it overcomes the critical issue of spectral bias, leading to highly accurate and physically consistent solutions. This architectural approach represents a promising direction for the future of machine learning, enabling the application of deep learning to a wider class of challenging problems in science and engineering.

This work contributes to the growing evidence that carefully designed neural network architectures, when properly integrated with physical priors, can overcome fundamental limitations of both traditional numerical methods and naive machine learning approaches. ASPEN demonstrates that spectral bias—long considered an inherent weakness of neural networks—can be transformed into a strength through adaptive learning of the spectral basis itself.

The implications extend beyond the specific PDEs studied here. Many scientific and engineering domains involve stiff, multi-scale, nonlinear dynamics: from pattern formation in biological systems \cite{Torabi2019PatternFIY}, to phase transitions in condensed matter \cite{Du1992AnalysisAAI}, to turbulence in fluid mechanics. ASPEN's mesh-free nature, differentiability, and ability to handle inverse problems position it as a valuable tool for these communities, particularly when traditional mesh-based methods become computationally prohibitive or when sparse experimental data must be integrated with physical models.

Looking forward, we envision ASPEN-like frameworks becoming part of a broader toolkit for \emph{physics-aware machine intelligence}, where domain knowledge and data-driven learning are synergistically combined to tackle problems at the frontier of scientific computing. The key insight—that inductive biases should not merely be accepted but can be adaptively optimized as part of the learning process—may inform future developments well beyond the specific context of physics-informed neural networks.

\section{Conclusion}
In this work, we addressed the challenge of applying physics-informed deep learning to stiff, nonlinear dynamical systems, a domain where standard architectures often fail. We demonstrated that a conventional PINN, built from a standard MLP, is fundamentally incapable of solving the complex Ginzburg-Landau equation. This baseline model, plagued by spectral bias, failed to represent the solution's high-frequency and multi-scale features, leading to a catastrophic divergence from the correct physical dynamics. This failure highlighted the critical need for novel architectures that can explicitly manage and overcome this inductive bias. To address this, we proposed the Adaptive Spectral Physics-Enabled Network (ASPEN), a framework that integrates an adaptive spectral input layer directly into the network. By treating the frequencies of this Fourier feature mapping as learnable parameters, ASPEN dynamically tunes its own spectral basis during training. This mechanism allows the model to efficiently allocate its representational power to the specific frequencies required by the PDE's solution, effectively bypassing the spectral bias that cripples standard MLPs. Our numerical experiments provided a clear validation of ASPEN's capabilities. In contrast to the baseline, the ASPEN mode successfully solved the CGLE with high fidelity, producing a solution visually indistinguishable from the high-resolution ground truth. This qualitative success was substantiated by strong quantitative metrics: the model converged to a low median physics residual of just $5.10 \times 10^{-3}$, demonstrating the PDE was satisfied across the domain. Detailed analysis of solution slices at various times confirmed this precision, showing a near-perfect overlay with the ground truth and a maximum domain wall position error of only $0.11$ at the final time $t=10.0s$. Most critically, ASPEN proved to be not just accurate but physically consistent, precisely tracking emergent physical properties like the system's free energy relaxation and the stable equilibrium of the domain wall-complex behaviors the baseline modl failed to capture. Ultimately, the contribution of this work is a robust and effective framework for solving complex, nonlinear PDEs that remain intractable for standard PINNs. We have shown that by integrating the network with a learnable, adaptive spectral basis, we can create a solver that is accurate, stable, and physically consistent.

\section*{Acknowledgments}
This work was conducted under the supervision and research governance of the Lenggoro Laboratory, Tokyo University of Agriculture and Technology (TUAT). The first author was supported by the TUAT Special Program Scholarship and a Research Assistantship from the Lenggoro Laboratory during Oct 2025–Mar 2026, and by a Lenggoro Laboratory Research Assistantship from Apr 2026 onward. The authors acknowledge TUAT/Lenggoro Lab computing resources, including a workstation with Intel Core i9-12900K and NVIDIA RTX A4000 (16 GB).

\bibliographystyle{unsrt}

\end{document}